\documentclass[manuscript,screen,nonacm, authorversion]{acmart}
\usepackage{geometry}
\renewcommand\footnotetextcopyrightpermission[1]{} %
\setcopyright{none}

\AtBeginDocument{%
  \providecommand\BibTeX{{%
    \normalfont B\kern-0.5em{\scshape i\kern-0.25em b}\kern-0.8em\TeX}}}

\usepackage[inline]{enumitem}
\usepackage{changepage}
\usepackage{caption}
\usepackage{subcaption}
\usepackage[utf8x]{inputenc} 
\usepackage{tikz}
\usepackage{thm-restate}
\usepackage{mathtools}
\usepackage[capitalise]{cleveref}
\usetikzlibrary{positioning}

\newtheoremstyle{my_thms}%
  {.5\baselineskip\@plus.2\baselineskip\@minus.2\baselineskip}%
  {.5\baselineskip\@plus.2\baselineskip\@minus.2\baselineskip}%
  {\normalfont}%
  {}%
  {\itshape}%
  {.}%
  {.5em}%
  {\thmname{\textbf{\textit{#1}}}\thmnumber{ {\textbf{\textit{#2}}}}\thmnote{ (#3)}}

\theoremstyle{my_thms}

\newtheorem{theorem}{Theorem}[section]

\newtheorem{proposition}[theorem]{Proposition}
\newtheorem{lemma}[theorem]{Lemma}

\newtheorem{definition}[theorem]{Definition}

\theoremstyle{acmplain}

\SetEnumitemKey{ncases}{itemindent=!,before=}

\newcommand{\prfcase}[2]{
\vspace{1em}
\textit{#1:} #2} 

\newcommand{\StrIndCase}[2]{
\vspace{3em}
\textit{Case:} #2 \hspace*{\fill} \textit{(#1)}}

\newenvironment{iproof}{%
\pushQED{\qed}
\noindent\textsc{Proof.}
\setlength{\parindent}{0em}
}{
\popQED
\vspace{1em}
}

\newenvironment{iproof_of}[1]{%
\vspace{-0.5em}
\pushQED{\qed}
\noindent
{\textsc{Proof of \ref{#1}.}}
\setlength{\parindent}{0em}
}{
\popQED
\vspace{1em}
}

\newenvironment{proofsketch}[1]{
\pushQED{\hspace*{\fill}\textit{(Full proof in \cref{prf:#1})}}
\noindent\textsc{Proof Overview.} 
}{
\popQED
}

\newenvironment{StrProofSketch}[1]{
\pushQED{\hspace*{\fill}\textit{(Further Cases in \cref{prf:#1})}}
\noindent\textsc{Proof Overview.} \textit{(Full Proof in \cref{prf:#1})} 
}{
\popQED
}

\newenvironment{subproof}{

\pushQED{\qed}
\begin{adjustwidth}{38pt}{0em}
\hspace{-38pt}{\textit{Subproof.}} 
}{
\popQED
\end{adjustwidth}
}

\newcommand{\StrIndCaseSketch}[2]{
\vspace{0.5em}
\textit{Case:} #2 \hspace*{\fill} \textit{(#1)}} 

\newenvironment{subproofsketch}{

\pushQED{\qed}
\begin{adjustwidth}{52pt}{0em}
\hspace{-42pt}\textit{Subproof.} 
\setlength{\parindent}{0em}
}{
\popQED
\end{adjustwidth}
}

\def\ConvColor{blue!40}
\def\SumColor{rgb:blue,4;green,8}

\def\NonInvColor{red!100}
\def\HNColor{green!70}
\def\StdColor{black}
\def\FHNColor{blue!60}

\def\FMColor{orange}

\DeclareRobustCommand\purplebox{\tikz{\draw[fill=\ConvColor, opacity=0.8, draw=none] (6.69,0) rectangle ++(0.15,0.15);}}

\DeclareRobustCommand\greenbox{\tikz{\draw[fill=\SumColor, opacity=0.7, draw=none] (6.69,0) rectangle ++(0.15,0.15);}}

\newcommand{\hood}{\ensuremath{\mathbb{B}^\circ}}
\newcommand{\ball}[2]{\ensuremath{\mathbb{B}_{#1}(#2)}}
\newcommand{\balls}{\ensuremath{\mathcal{B}}}
\newcommand{\bradius}[1]{\ensuremath{\mathcal{R}(#1)}}
\newcommand{\bcenter}[1]{\ensuremath{\mathcal{C}(#1)}}
\newcommand{\relint}{\ensuremath{\operatorname{relint}}}
\newcommand{\aff}{\ensuremath{\operatorname{aff}}}

\newcommand{\bHull}{\ensuremath{\mathcal{H}_\infty}}
\newcommand{\ldot}{\mathpunct{.}}
\newcommand{\st}{\mathpunct{.}}
\newcommand{\wehave}{\mathpunct{.}}
\newcommand{\from}{\colon}
\newcommand{\given}{\colon}
\newcommand{\reals}{\ensuremath{\mathbb{R}}}
\newcommand{\nats}{\ensuremath{\mathbb{N}}}
\newcommand{\relu}{\ensuremath{\operatorname{ReLU}}}
\newcommand{\sign}{\ensuremath{\operatorname{sign}}}

\newcommand{\flipX}{\ensuremath{\hat{\mathtt{x}}}}
\newcommand{\flipL}{\ensuremath{\hat{\mathtt{l}}}}
\newcommand{\defeq}{\ensuremath{\coloneqq}}

\newcommand{\boxhull}{$l_\infty$-hull}

\renewcommand{\citet}[1]{\cite{#1}}

\begin{document}

\title{The Fundamental Limits of Interval Arithmetic for Neural Networks}

\author{Matthew Mirman}
\email{matthew.mirman@inf.ethz.ch}
\orcid{0000-0002-2109-3272}
\affiliation{%
  \institution{ETH Zurich}
  \country{Switzerland}
}

\author{Maximilian Baader}
\email{mbaader@inf.ethz.ch}
\orcid{0000-0002-9271-6422}
 \affiliation{%
   \institution{ETH Zurich}
   \country{Switzerland}
 }

\author{Martin Vechev}
\email{martin.vechev@inf.ethz.ch}
\affiliation{%
  \institution{ETH Zurich}
  \streetaddress{Ramistrasse 101}
  \postcode{8092}
  \city{Zurich}
  \country{Switzerland}
}

\renewcommand{\shortauthors}{Mirman, et al.}

\begin{abstract}
Interval analysis (or interval bound propagation, IBP) is a popular technique for verifying and training provably robust deep neural networks, a fundamental challenge in the area of reliable machine learning. However, despite substantial efforts, progress on addressing this key challenge has stagnated, calling into question whether interval arithmetic is a viable path forward.

In this paper we present two fundamental results on the limitations of interval arithmetic for analyzing neural networks. Our main impossibility theorem states that for any neural network classifying just three points, there is a valid specification over these points that interval analysis can not prove. Further, in the restricted case of one-hidden-layer neural networks we show a stronger impossibility result: given any radius $\alpha < 1$, there is a set of $O(\alpha^{-1})$ points with robust radius $\alpha$, separated by distance $2$, that no one-hidden-layer network can be proven to classify robustly via interval analysis.

\end{abstract}

\maketitle

\begin{figure}[t]
    \centering
    \begin{subfigure}[b]{0.6\textwidth}
        \centering
        \begin{tikzpicture}[scale=0.9]
    \pgfmathsetmacro{\l}{1.3}
    \pgfmathsetmacro{\w}{3}

    \definecolor{mypurple}{HTML}{e2ddff}

    \node[] (X) at (- 0.5 * \w, - 0.5 * \l) {$x$}; 

    \node[shape=circle,draw=gray, fill=mypurple, minimum size=20pt] (A) at (0,0) {$0$};
    \node[shape=circle,draw=gray, fill=mypurple, minimum size=20pt] (B) at (0,-\l) {$0$};

    \node[shape=circle,draw=gray, fill=mypurple, minimum size=20pt] (C) at (\w,0) {$2$};
    \node[shape=circle,draw=gray, fill=mypurple, minimum size=20pt] (D) at (\w,-\l) {$-2$};

    \node[shape=circle,draw=gray, fill=mypurple, minimum size=20pt] (E) at (2 * \w,0) {$0$};
    \node[shape=circle,draw=gray, fill=mypurple, minimum size=20pt] (F) at (2 * \w,-\l) {$1$};

    \node[] (FX) at (2 * \w + 0.5 * \w, - 0.5 * \l) {$f(x)$};

    \path [->] (X) edge node[left,above] {$1$} (A);
    \path [->] (X) edge node[left,below] {$-1$} (B);

    \path [->] (A) edge node[left,above] {$-1$} (C);
    \path [->] (B) edge node[left,below] {$1$} (D);
    \path [->] (A) edge node[pos=0.4,left,above] {$1$} (D);
    \path [->] (B) edge node[pos=0.4,left,below] {$-1$} (C);
    
    \path [->] (C) edge node[left,above] {$1$} (E);
    \path [->] (D) edge node[left,below] {$-1$} (F);
    \path [->] (C) edge node[pos=0.4,left,above] {$0$} (F);
    \path [->] (D) edge node[pos=0.4,left,below] {$0$} (E);

    \path [->] (E) edge node[right,above] {$1$} (FX.north west);
    \path [->] (F) edge node[right,below] {$-1$} (FX.south west);

\end{tikzpicture}
        \vspace{-0.5em}
        \caption{Network. Encircled numbers are biases, and circles (\purplebox) are \relu{} activations.}
        \vspace{1em}
    \end{subfigure}
    \hfill
    \begin{subfigure}[b]{0.35\textwidth}
        \centering
        \includegraphics[width=\textwidth]{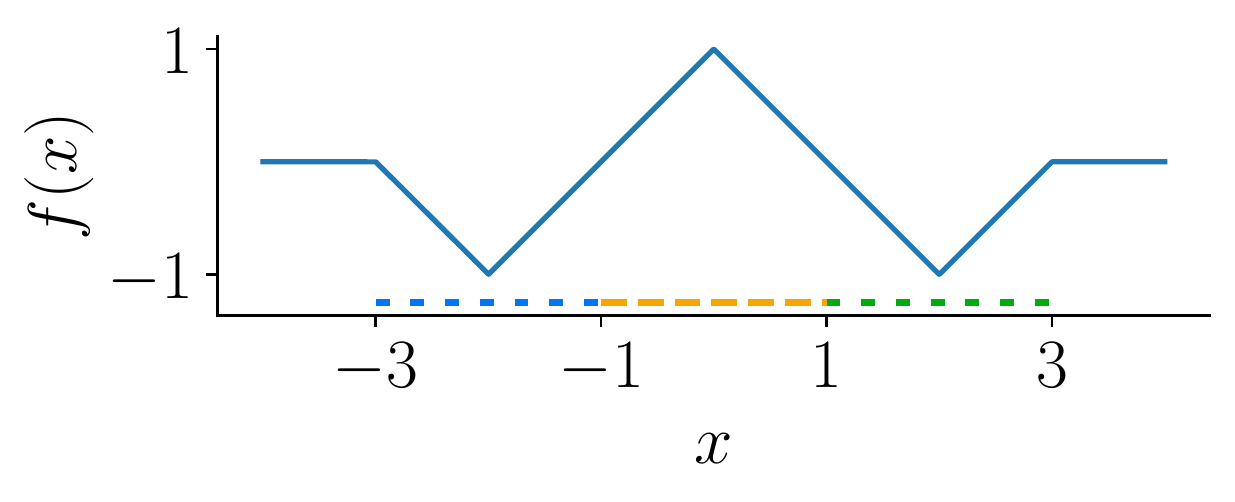}
        \vspace{-2em}
        \caption{Network values $f(x)$ for $x \in [-4, 4]$.}
        \vspace{1em}
        \label{fig:network:values}
    \end{subfigure}
    \begin{subfigure}[b]{0.14\textwidth}
        \centering
        \includegraphics[width=\textwidth]{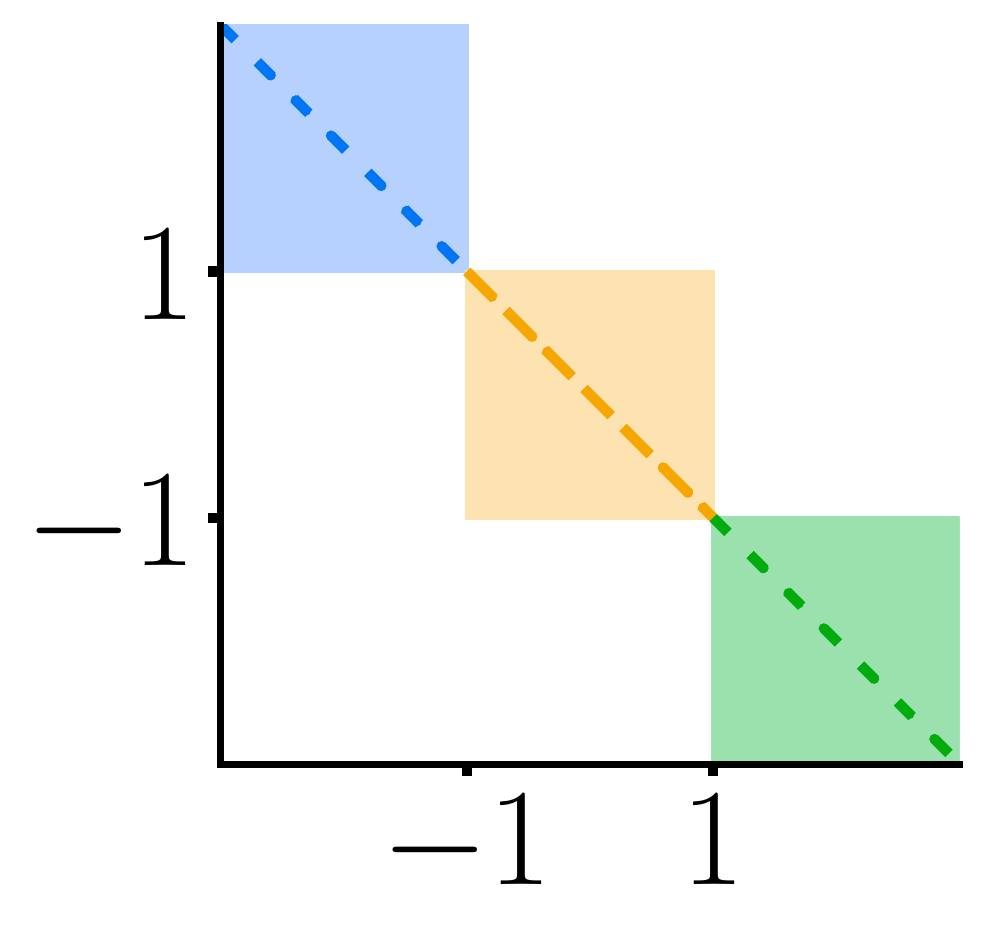}
        \caption{1st affine layer}
    \end{subfigure}
    \hfill
    \begin{subfigure}[b]{0.14\textwidth}
        \centering
        \includegraphics[width=\textwidth]{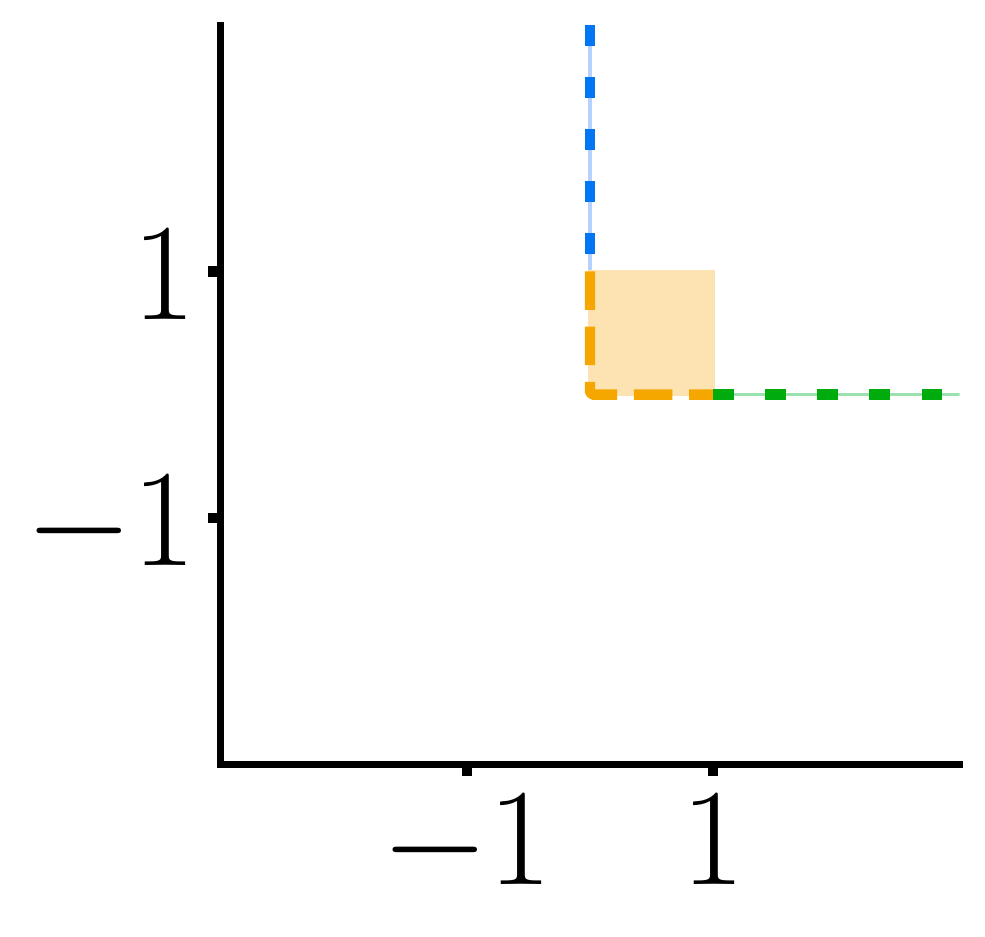}
        \caption{1st ReLU layer}
    \end{subfigure}
    \hfill
    \begin{subfigure}[b]{0.14\textwidth}
        \centering
        \includegraphics[width=\textwidth]{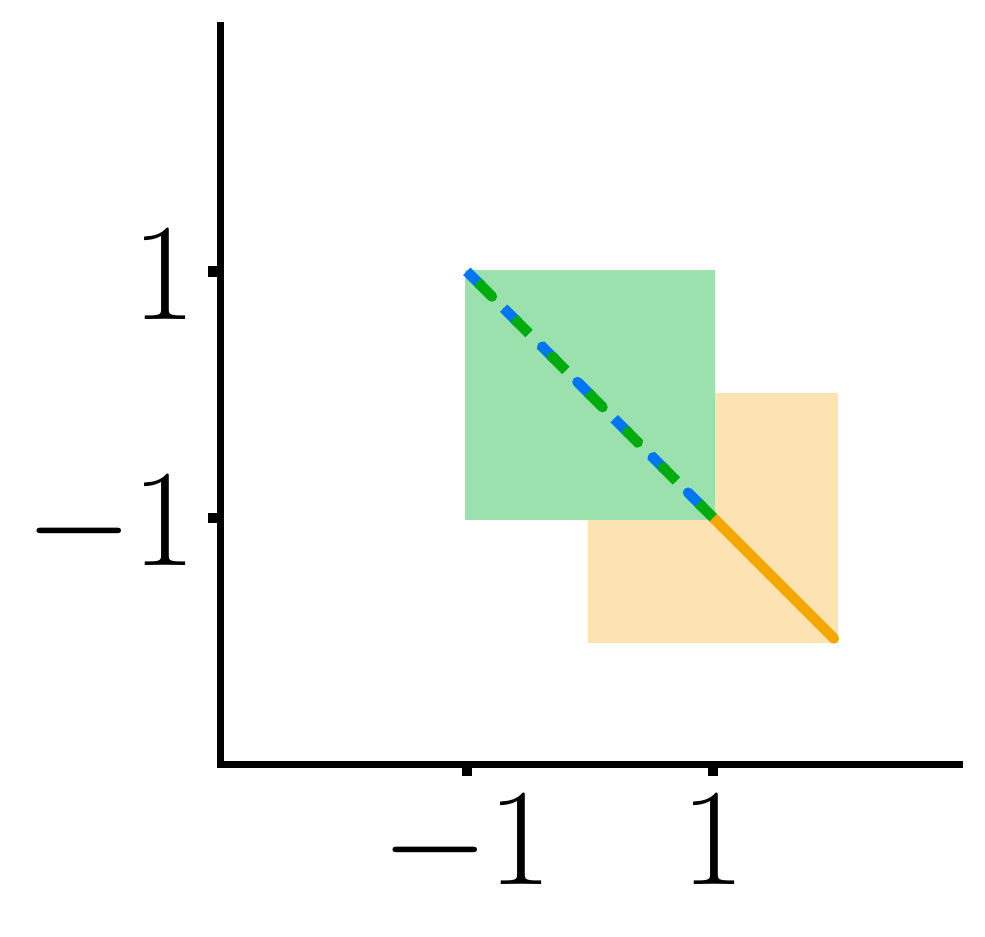}
        \caption{2nd affine layer}
        \label{fig:network:2affine}
    \end{subfigure}
    \hfill
    \begin{subfigure}[b]{0.14\textwidth}
        \centering
        \includegraphics[width=\textwidth]{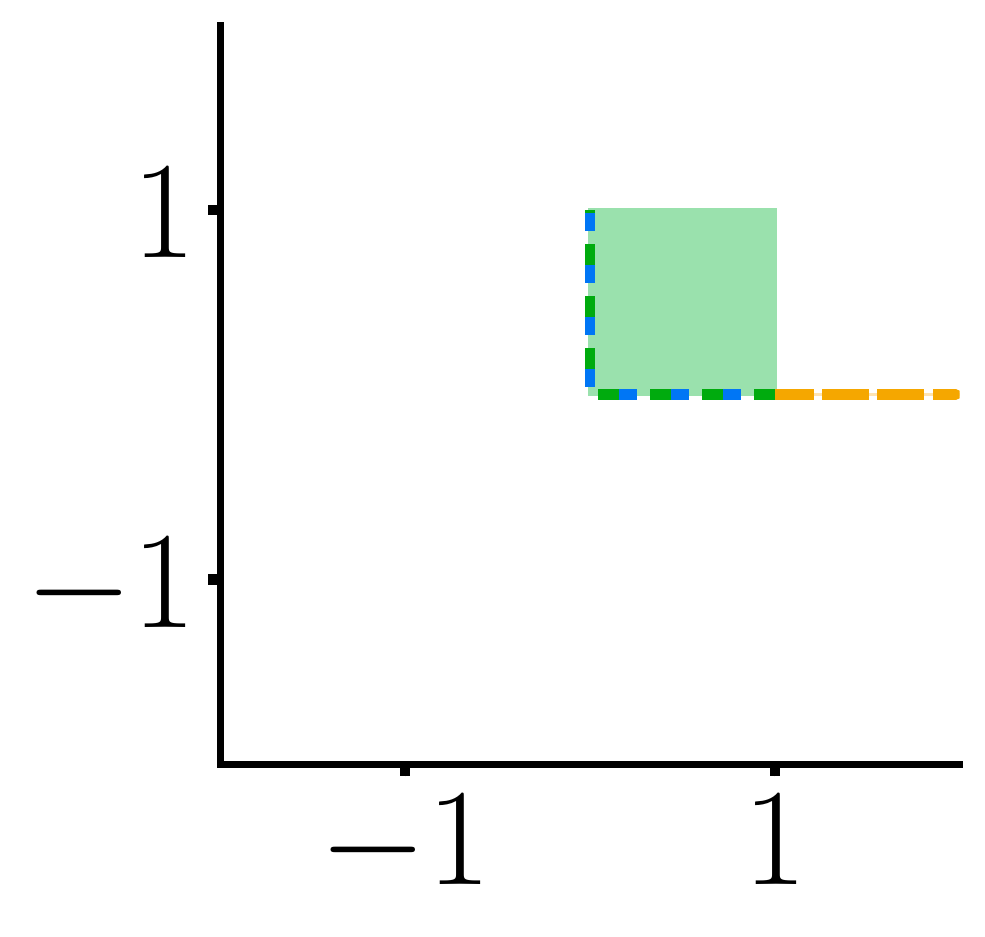}
        \caption{2nd ReLU layer}
    \end{subfigure}
    \hfill
    \begin{subfigure}[b]{0.14\textwidth}
        \centering
        \includegraphics[width=\textwidth]{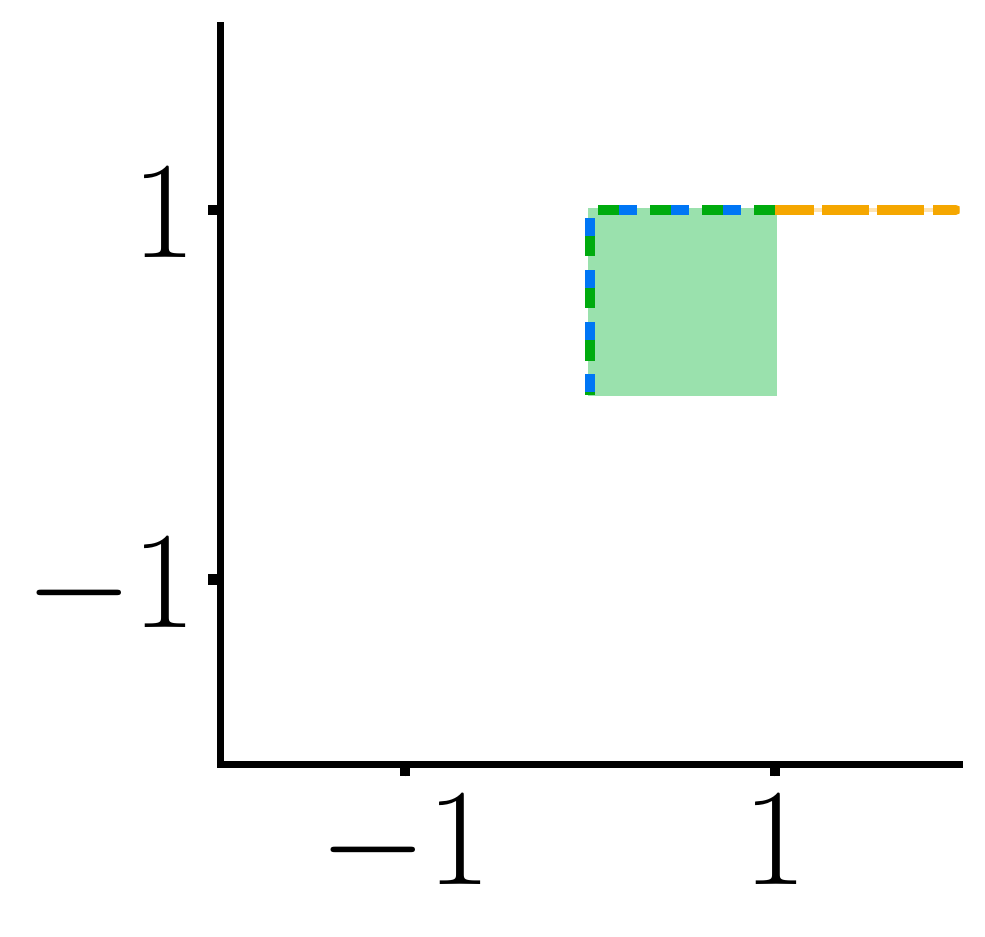}
        \caption{3rd affine layer}
    \end{subfigure}
    \hfill
    \begin{subfigure}[b]{0.14\textwidth}
        \centering
        \includegraphics[width=\textwidth]{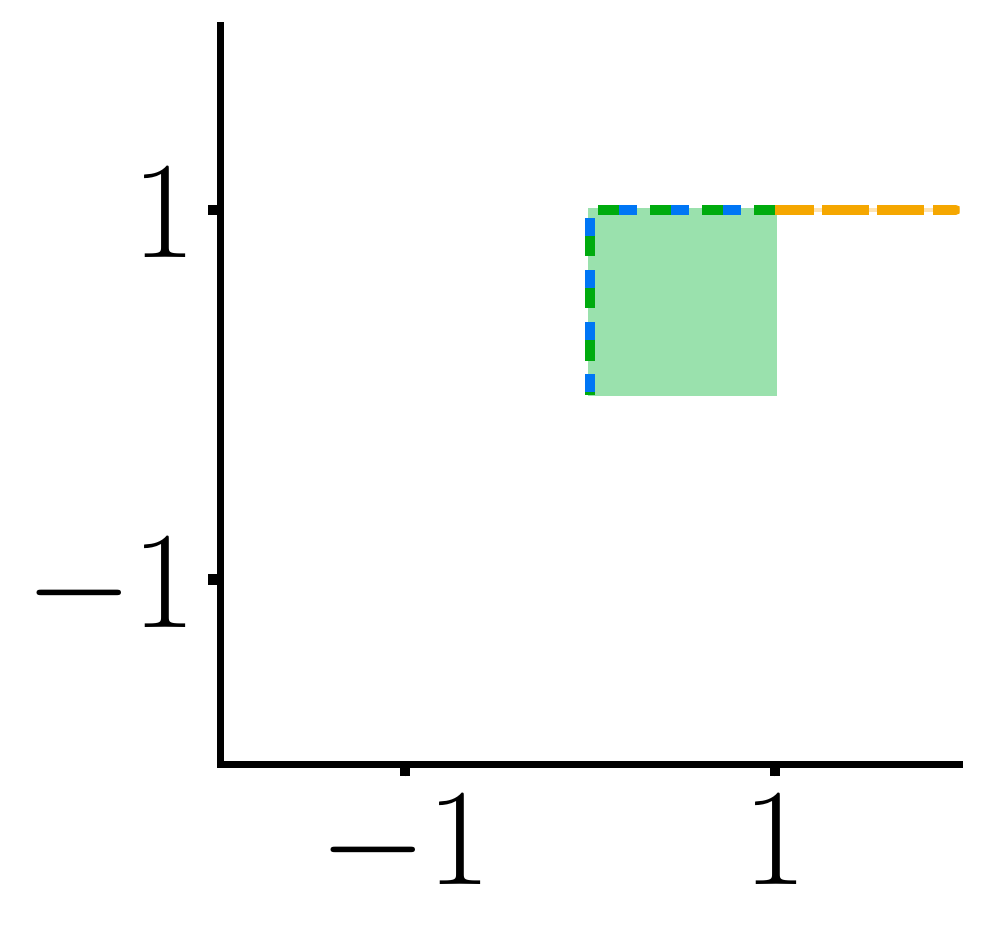}
        \caption{3rd ReLU layer}
    \end{subfigure}
    \definecolor{myblue}{HTML}{0076F5}
    \definecolor{myorange}{HTML}{F5A700}
    \definecolor{mygreen}{HTML}{00AB0E}
    \DeclareRobustCommand\mybluebox{\tikz{\draw[fill=myblue, opacity=0.8, draw=none] (6.69,0) rectangle ++(0.15,0.15);}}
    \DeclareRobustCommand\myorangebox{\tikz{\draw[fill=myorange, opacity=0.8, draw=none] (6.69,0) rectangle ++(0.15,0.15);}}
    \DeclareRobustCommand\mygreenbox{\tikz{\draw[fill=mygreen, opacity=0.8, draw=none] (6.69,0) rectangle ++(0.15,0.15);}}
    \caption{An example of a neural network which is in fact robust, yet which interval arithmetic fails to {\em prove} is robust. The three intervals 
    $L=[-3, -1]$ (\mybluebox), 
    $M=[-1, 1]$ (\myorangebox) and 
    $R=[1, 3]$ (\mygreenbox) are depicted using dashed lines in \cref{fig:network:values}. The interval propagation through the network is shown using the rectangles, the concrete values are shown by the dashed lines. Here, after the 2nd affine layer (\cref{fig:network:2affine}) the orange and green box overlap although the orange lines do not overlap, showing the loss of precision. The output Interval is $[-1, 1]$ for all three input intervals $L, M$ and $R$.}
\vspace{-1em}
    \label{fig:network}
\end{figure}
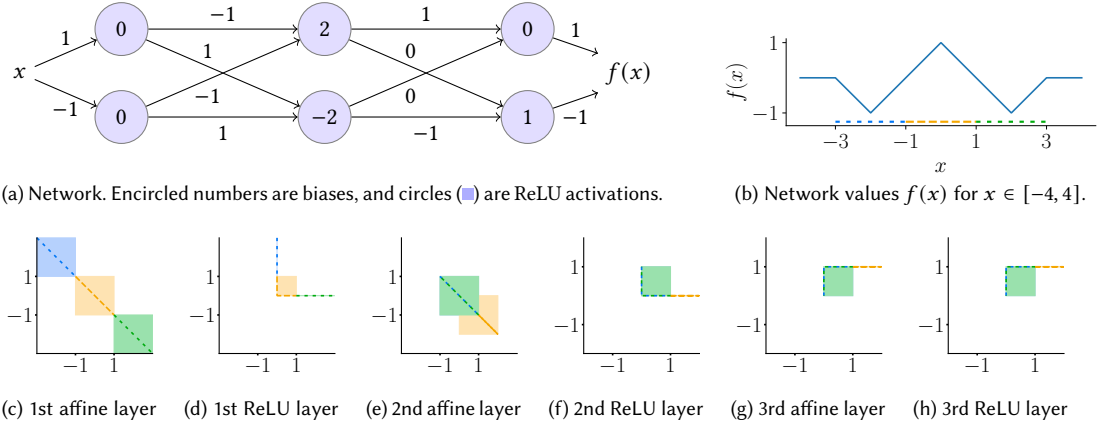

\section{Introduction}

As neural networks are increasingly used in safety critical environments, ensuring their behavior with {\em formal verification} has become a highly active research direction~\cite{liu2019algorithms,huang2020survey}. Because neural networks are often too large for complete verification methods, incomplete analysis techniques are frequently employed~\cite{ai2} -- these can scale to larger models though may fail to prove a property that actually holds (as demonstrated in \cref{fig:network}). Indeed, recent progress in constructing provable neural networks has been achieved thanks to leveraging incomplete methods, and particularly interval (box) bound propagation~\cite{diffai}. However, while many improvements to {\em provable defenses} have been published~\cite{Gowal2018, zhang2018crown, zhang2020towards, xiao2019training, Wong2018, liu2021training, boopathy2021fast, shi2021fast, xu2021fast} (most building on IBP), progress remains far from satisfactory: the state-of-the-art certified robust accuracy is roughly 60\% on CIFAR10 \cite{balunovic2019adversarial}, compared to state-of-the-art standard accuracies of above 95\%.
The stagnation of progress in constructing provably robust neural networks, and the importance of interval arithmetic to other areas of mathematics \cite{tucker2002rigorous}, has led to a fundamental question:

\vspace{0.75em}
\centerline{Do neural networks exist which can be efficiently (with interval analysis) proven correct?}
\vspace{-0.2em}
\centerline{\bf \footnotesize (Fundamental Theoretical Question) }
\vspace{0.75em}

The first result addressing this question was investigated in \citet{baader2020universal} which proved an analog to the universal approximation theorem \cite{cybenko1989approximation, hornik1989multilayer} for interval-analyzable networks.  \citet{wang2020interval} further showed that two hidden layer networks could also be interval-analyzable approximators. \cite{anonymous2022on} also demonstrated that training with interval propagation converges with high probability.
While it is helpful to know that searching for networks which can be easily analyzed might not be futile,
these results do not explain, and even contradict the provable training gap that is observed in practice.
A preliminary negative result was shown in \cite{wang2020interval}: verifying the robustness of arbitrary neural networks in general and thus translating arbitrary neural networks into interval-analyzable forms is NP-hard.
In our work, we provide a strong negative answer, thus explaining the provable training gap: we demonstrate that non-trivial datasets can not be classified by interval-provable networks.

\newpage
Formally, given a neural network, or more generally any program, $f \from \mathcal{X} \to \mathcal{Y}$,
the goal of verification is to algorithmically prove that $f$ maps an
input specification, $S_I \subseteq \mathcal{X}$, to a subset of an output specification, $S_O \subseteq \mathcal{Y}$, where $(S_I,S_O)$ is a member of a set $\mathcal{S} \subseteq \mathcal{P}(\mathcal{X}) \times \mathcal{P}(\mathcal{Y})$, which we call the {\em specification task}.
Interval analysis in particular replaces the basic operations of $f$
with interval arithmetic~\cite{moore1996interval, hickey2001interval}, producing a {\em sound} interval extension, $f^\# \from \text{Intervals}(\mathcal{X}) \to \text{Intervals}(\mathcal{Y})$, of $f$ such that every element of $S_I$
is mapped by $f$ to an element of $f^\#(S_I)$.
As representing and computing intervals is efficient, $f$ is proven to meet the specification $(S_I,S_O)$
by proxy of proving $f^\#(\mathcal{S}_I) \subseteq S_O$.
The specification tasks we consider are {\em $l_\infty$ robust classifications},
meaning the input specifications are closed $l_\infty$-balls (i.e., intervals), and output specifications are either $\reals_{>0}$ or $\reals_{<0}$.  Formally, we say an $l_\infty$-robustness classification $\mathcal{S}$ is {\em complete},
if for any specification $(S_I,S_O) \in \mathcal{S}$,
there is an $l_\infty$-ball, the {\em scheme},  $T_S \supseteq S_I$ such that for all $l_\infty$-balls $B \subseteq T_S^\circ$, we have $(B,S_O) \in \mathcal{S}$.

\vspace{1em}
\noindent{\bf Main contributions.}
In this paper, we present the first proofs capturing key limitations (incompleteness) of interval analysis for neural networks:
\begin{itemize}
\item \textbf{General interval impossibility (\textbf{\cref{thm:ex}})}: It is impossible to construct a feed-forward \relu-neural network of any shape (e.g., residual, convolutional, dense, fully-connected) that is {\em completely provably robust} (\cref{def:complete}) with interval analysis for a simple one-dimensional dataset with only three points.
\vspace{0.25em}
\item \textbf{One-layer strong interval impossibility (\cref{thm:ub})}: Even when the requirement for complete provability is relaxed to regions that are distant from each other ($\alpha$-interval provable with $\alpha<1$ as in \cref{def:plain_classifier}), there are datasets with $O(\alpha^{-1})$ points that can not be provably robustly classified with one-hidden layer networks using interval analysis.
\vspace{0.25em}
\item \textbf{One-layer strong interval-agnostic possibility (\cref{prop:existance})}:
{\em completely-robust classifiers} can always be constructed with one-hidden layer networks, even if they are not necessarily provably robust using interval analysis.   Together with \cref{thm:ub} and \cref{thm:ex} this implies that the restriction that a network be  analyzable with interval-arithmetic is severely limiting.
\end{itemize}

\section{Problem Motivation}

Studying the robustness of artificial neural networks has become an important area of research, as neural networks are increasingly deployed in safety-critical applications such as self-driving cars
~\cite{BojarskiTDFFGJM16}. 
\citet{adversarialDiscovery} first demonstrated that neural networks classifying images can be fooled into misclassification by imperceptible pixel perturbations in an otherwise correctly classified image. 

Many of these fooling techniques, known as adversarial attacks, have been developed ~\cite{
Carlini017,
Goodfellow2015,
KurakinGB16,
ShahamYN15,
croce2019sparse,
papernot2016transferability,
wong2019wasserstein}.
To defend against these attack, methods hardening models have been proposed
~\cite{PapernotMWJS15,tramer2017ensemble, wong2019fast, stutz2020confidence, BastaniILVNC16, croce2020reliable}.
A particular line of research aims to provide {\em formal} guarantees 
(i.e., verify) that neural networks behave correctly~\cite{katz2017reluplex, Singh2018,Singh2019,Boopathy2019,Liu2019,Wang2018a,Balunovic2019,zhang2021certified,lin2020certified,croce2019provable,Croce2018}.
As complete verification of a neural network is NP-Hard~\cite{katz2017reluplex}, the majority 
of modern techniques is incomplete and are based on over-approximating the behavior of a network~\cite{ai2}.
While incomplete methods can be highly efficient, it a correct classification of a network {\em might not be provably} correct, as is illustrated in \cref{fig:network}. 
In fact, for naturally trained neural networks, only a small percentage of non-attackable input images are verifiable. 

To improve verification rates, techniques to training networks that are amenable to verification~\cite{Raghunathan2018,diffai,kolter2018provable,Wong2018} have been developed.
While this has been a very active area of research, the state-of-the-art developed by \citet{balunovic2019adversarial},
achieves a certified robust accuracy of 60.5\% on CIFAR10 which is unsatisfactory compared to a state-of-the-art standard accuracy of above 95\%.

The recent plateau of progress in closing this gap has raised concerns about whether there are theoretical limitations to neural network analysis \cite{Salman2019}.
In this work, we provide fundamental limitations, which helps to explain the significant gab between certified robust accuracy, and standard accuracy. 
We focus on interval analysis, as some of the most successful and widely used methods have been based on it~\cite{diffai, Gowal2018}.

\section{Background}

In this section, we introduce the main concepts, and notation central to understanding our results.

\subsection{General Notation}

The main results in this work centers around the interval domain, which is technically the domain of axis-aligned 
bounding boxes, and thus we begin by describing notation related to such boxes.

Let $\balls^d$, the set of closed, non-empty, axis-aligned boxes
of dimension $d$ (also known as $l_\infty$ balls). 
We write the box with center $c \in \reals^d$ and radius $r\in \reals_{\ge 0}^d$ as
$\ball{r}{c} \defeq \{ x \given \forall i \in [d] \ldot \exists \xi \in [-1,1] \ldot x_i = c_i + \xi r_i  \}$.
For a given box $B \in \balls^d$ let $\bcenter{B}$ denote its center and $\bradius{B}$ 
denote its radius such that $B = \ball{\bradius{B}}{\bcenter{B}}$.
We also write $\hood_\epsilon(x) \defeq \{ y \in \reals^d \given || x - y ||_\infty < \epsilon \}$ for $x\in \reals^d$ and $\epsilon \in \reals_{>0}$.

For a set $Y$, we write $Y|_i$ for the restriction of $Y$ to the dimension $i$, or more formally, 
$Y|_i \defeq \{ y_i \given y \in Y \}.$
For any bounded and non-empty set $C$, the {\em \boxhull}, written $\bHull(C)$, 
is the smallest axis aligned box containing $C$.  
Formally, $\bHull(C) \defeq \{ x \given \forall i \in [d] \ldot \sup(C|_i) \leq x_i \leq \inf(C|_i)\}.$

If $f \from A \to B$ and $S \subseteq A$ we write $f[S] \defeq \{ f(s) \given s \in S \}$, but sometimes abuse notation and write $f(S) \defeq f[S]$ to avoid clutter.  Similarly, we also occasionally write $f^{-1} \circ g^{-1}$ even when $f$ and $g$ 
are non invertible to mean $(g\circ f)^{-1}$.  For any set $S$ we write $\mathcal{P}(S)$ to mean the powerset of $S$.
For some positive natural $k \in \nats$ we write $[k] \defeq \{1,\ldots,k\}$.

\subsection{Robustness and Interval Certification (IBP)}\vspace{0.5em} 

Suppose $f \from \reals^d \to \reals$ is some function (i.e., neural network).  
We say that this network assigns a label $l \in \{-1,1\}$ to a point $x \in \reals^d$ if $\sign f(x) = l$.
In our case, we discuss $l_\infty$-adversarial region specifications.  In this case, we say that $f$ is {\em $\epsilon$-robust around $x$} 
with {\em label} $l$ if $\forall x' \in \ball{\epsilon}{x} \wehave f(x') = l$.

The goal of robustness certification is to provide a guarantee that a neural network is robust at some point. 
However, robustness certification does not need to inform when a neural network is {\em not}-robust at a point.
This leads to efficient methods in terms of {\em over-approximation}, originally described as {\em abstract-interpretation}~\cite{CC77} and applied to neural networks by \citet{ai2}. 

\vspace{0.1em} 
\begin{definition}
\label{def:ai}
Given the {\em concrete-domains} $\mathcal{D}$ and $\mathcal{G}$ we say the 
{\em concrete function} $f \from \mathcal{D} \to \mathcal{G}$ {\em over-approximated} by the 
{\em transformed function (abstract transformer)} $f^\# \from \tilde{\mathcal{D}} \to \tilde{\mathcal{G}}$ with 
{\em abstract-domains}
$\tilde{\mathcal{D}} \subseteq \mathcal{P}(\mathcal{D})$ and $\tilde{\mathcal{G}} \subseteq \mathcal{P}(\mathcal{G})$
if it is {\em sound} if for any abstract set  $\mathcal{S} \in \mathcal{D}$ we have $f(\mathcal{S}) \subseteq f^\#(\mathcal{S})$
\end{definition}
\vspace{0.1em} 

The goal of such an over-approximation that it is possible to computationally represent and modify elements of {\em abstract-domains}, whereas it is not in general possible to do this for any subset of $\mathcal{D}$ (as it might very well be $\reals$).
We note our definitions are a slight departure from the traditional abstract interpretation literature.  
While typically, the abstract domain refers to the set of {\em representations} of subsets of the concrete domain, 
and the abstract transformer acts on these representations, we refer to the sets they represent themselves.
As here we are concerned only with the question of the mathematical limitations of what is represented, and not the question of how to efficiently represent 
or perform computations (this is trivial for IBP), we can simplify our presentation dramatically by discussing only the represented sets and not the representations themselves.

To certify a function $f = g_n \circ \cdots \circ g_1$, where $g_i \from \reals^{d_i} \to \reals^{d_{i+1}}$, is $\epsilon$-robust at $x$ using abstract interpretation, one may pick abstract-domains $\mathcal{D}_i \subseteq \mathcal{P}(\reals^{d_i})$
and compute sound transformed functions, $g^\#_1, \ldots, g^\#_n$: If $g^\#_{i}$ and $g^\#_{i+1}$ over-approximates $g_i$ and $g_{i+1}$ 
then $g^\#_{i} \circ g^\#_{i+1}$ over-approximates $g_{i+1} \circ g_i$.   Then, given one can show for $f^\# = g^\#_n \circ \cdots \circ g^\#_1$, that for some $\tilde{A} \in \mathcal{D}_1$ such that $\ball{\epsilon}{x} \subseteq \tilde{A}$, it is true 
that $\forall \tilde{y} \in f^\#(\tilde{A}) \wehave \sign \tilde{y} = f(x)$, then one will have also shown that $\forall x' \in \ball{\epsilon}{x} \wehave \sign f(x') = f(x)$.

For any function $f \from A \to B$, we say that the {\em perfect transformation} of $f$ is $f^\mathcal{P} \from \mathcal{P}(A) \to \mathcal{P}(B)$ where for any set $S \subseteq A$ we have $f^\mathcal{P}(S) \defeq f[S]$.

While the perfect transformation of $f$ is just that, always perfect, it is important to note that over-approximation is typically {\em not precise}, meaning $f(S) \subsetneq f^\#(S)$.
In this case, it is in fact possible to not prove the guarantee, such as robustness, even if that guarantee holds for $f$.

\vspace{0.1em}
\paragraph{\bf Interval Analysis} In this paper, we focus on the {\em Interval (or Box)-domain}, and in particular, Interval Bound Propagation (IBP) \cite{Gowal2018}, which is also known as interval-analysis. In this case, $\balls^d$ is used as the abstract domain, $\tilde{\mathcal{D}}$, when the concrete domain is $\reals^d$.  An interval, $B\in \balls^d$, can either be represented as a center $c\in\reals^d$ and radius $r \in \reals^d_{\geq 0}$ as before, or as a lower-bound and upper bound, $\iota_l,\iota_u \in \reals^d$ respectively such that for each dimension $j\in[d]$, we have $\iota_{l,j} \leq \iota_{u,j}$.  
The two representations are related as follows: $\iota_l = c - r$ and $\iota_u = c + r$, 
or $c = \frac{1}{2}(\iota_l + \iota_u)$ and $\iota_u = \frac{1}{2}(\iota_u - \iota_l)$.

\vspace{0.1em}
\paragraph{\bf Analyzing Neural Networks} The application of interval analysis to neural networks with \relu-activations is straightforward.  In this paper, we consider (feed forward) neural networks defined inductively as follows:

\begin{definition}
\label{def:nets}
A $\sigma$-(neural) network with $\sigma$-activations, $f$, is any of the following forms:
\begin{itemize}
\item {\em Sequential Computation:} $f(x) = g_1(g_2(x))$ where $g_1$ and $g_2$ are also both $\sigma$-networks.
\item {\em Relational Duplication:} $f(x) = (x,x)$.
\item {\em Non-Relational Parallel Computation:} $f(x_1,x_2) = (g_1(x_1),g_2(x_2))$ where $g_1$ and $g_2$ are also both $\sigma$-networks.
\item {\em Constant:} $f(x) = \kappa$ for some constant $\kappa \in \reals$.
\item {\em Multiplication by a Constant:} $f(x) = \kappa \cdot x$ for some constant $\kappa \in \reals_{\neq 0}$.
\item {\em Activation:} $f(x) = \sigma(x)$.
\item {\em Relational Addition:} $f(x_1, x_2) = x_1 + x_2$.
\end{itemize}
\end{definition}

For the purposes of exploring its limits, we view IBP as method that implicitly constructs a transformed function which acts on intervals. We describe this transformed function inductively as well:

\begin{definition}[Interval Analysis]
\label{def:interval}
The {\em interval transformation}, $f^\#$, of a \relu-network $f$ is as follows:
\begin{itemize}
\item {\em Sequential Abstraction:} If $f(x) = g_1(g_2(x))$ then $f^\#(B) \defeq g^\#_1(g^\#_2(B))$.
\item {\em Relational Duplication:} If $f(x) = (x,x)$ then $f^\#(B) \defeq B \times B$.
\item {\em Non-Relational Parallel Abstraction:} 
If $f(x_1,x_2) = (g_1(x_1),g_2(x_2))$ then $f^\#(B) \defeq g^\#_1(B|_1) \times g^\#_2(B|_2)$.
\item {\em Constant:} If $f(x) = \kappa$ for $\kappa \in \reals$ where $m=n=1$, 
then $f^\#(B) \defeq \{ \kappa \}$.
\item {\em Multiplication by a Constant:} 
If $f(x) = \kappa \cdot x$ for some constant $\kappa \in \reals_{\neq 0}$ where $m=n=1$, 
then $f^\#(B) \defeq \{ \kappa \cdot x \}$.
\item {\em Activation:} 
If $f(x) = \relu(x)$ (where $m=n=1$), 
then $f^\#(B) \defeq \{  \relu(x) \given x \in B \}$.
\item {\em Relational Addition:} 
If $f(x_1, x_2) = x_1 + x_2$ where $m=2$ and $n=1$, 
then $f^\#(B) \defeq \{ a + b \given a \in B|_1 \wedge b \in B|_2 \}$.
\end{itemize}
\end{definition}

\begin{proposition} 
\label{prop:interval_sound}
If $f$ is a \relu-network then the interval transformer, $f^\#$, over-approximates $f$.
\end{proposition}

\section{Limits for Single Hidden Layer Networks}
\label{sec:single}

In this section we present an upper-bound on the number of points that can be proven to be  robustly classified with interval for a single-layer network.  
We do this by constructing a paradoxical dataset, which we call {\em flips}.
We begin by formalizing this dataset, and the notion of robust and provably robust on this dataset.

\begin{definition} We say:
\label{def:plain_classifier}
\begin{itemize}
\item A {\em flip} is a point $\flipX_i \defeq 2i$ with {\em label} $\flipL_i \defeq (-1)^i$.

\item $f \from \reals \to \reals$ is a {\em classifier for $k$ flips} if $\forall i \in [k] \wehave f(\flipX_i) = \flipL_i$.

\item $F \from \balls \to \mathcal{P}(\reals)$ is an {\em $\alpha$-classifier for $k$ flips} if 
$\forall i \in [k] \wehave \forall y \in F([\flipX_i -\alpha, \flipX_i +\alpha]) \wehave \sign y = \flipL_i \wedge F(\{\flipX_i\}) = \{\flipL_i\}$.

\item If $f^\mathcal{P}$ (the perfect transformation) 
is an $\alpha$-classifier for $k$ flips, 
we say $f$ is an {\em $\alpha$-robust classifier} for {\em $k$ flips}.

\item If $f^\#$ (the interval transformation from \cref{def:interval}) 
is an $\alpha$-classifier for $k$ flips, 
then we say that $f$ is a {\em provably $\alpha$-robust classifier} for {\em $k$ flips}.

\end{itemize}
\end{definition}

We now specify the notion of single-layer network for which we demonstrate bounds:

\begin{definition}
\label{def:single_layer_net}
A {\em single-layer $\sigma$-network}, $f \from \reals \to \reals$, with {\em $n$-neurons} and {\em $\sigma$-activations}
is a function
with {\em pre-activation weights}, $N \in \reals^n$, 
     {\em pre-activation bias}, $b \in \reals^n$, 
     {\em post-activation weights}, $M \in \reals^n$, 
 and {\em post-activation bias}, $d \in \reals$ (the weights and biases are known as {\em parameters}),
such that $f(x) = M \cdot \sigma(Nx + b) + d$.
\end{definition}

We note that while the \cref{def:interval} defines an ordering of addition, this 
definition does not.  While concrete-addition is associative, abstract addition is not always. However, thankfully, for the interval transformation, it is, and the bounds we demonstrate apply to any ordering of the operations. %

\newcommand{\sA}[3]{\ensuremath{\mathcal{A}_{#1,#2}(#3)}}
\newcommand{\sP}[3]{\ensuremath{\mathcal{I}_{#1,#2}(#3)}}

\begin{figure}
    \centering
    \vspace{-0.5em}
    \begin{subfigure}[b]{.14\textwidth}
        \centering
        \includegraphics[width=0.8\linewidth]{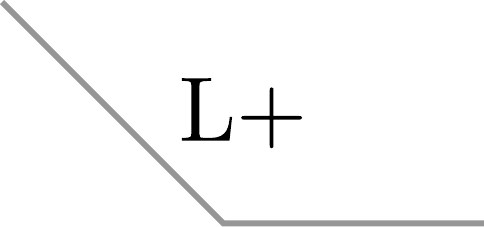}
        \caption{$\relu(-x)$}
    \end{subfigure}
    \hspace{1cm}
    \begin{subfigure}[b]{.14\textwidth}
        \centering
        \includegraphics[width=0.8\linewidth]{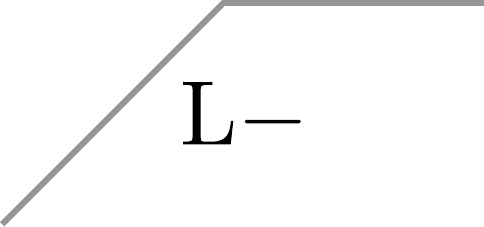}
        \caption{$-\relu(-x)$}
    \end{subfigure}
    \hspace{1cm}
    \begin{subfigure}[b]{.14\textwidth}
        \centering
        \includegraphics[width=0.8\linewidth]{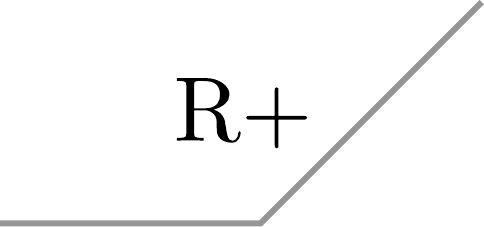}
        \caption{$\relu(x)$}
    \end{subfigure}
    \hspace{1cm}
    \begin{subfigure}[b]{.14\textwidth}
        \centering
        \includegraphics[width=0.8\linewidth]{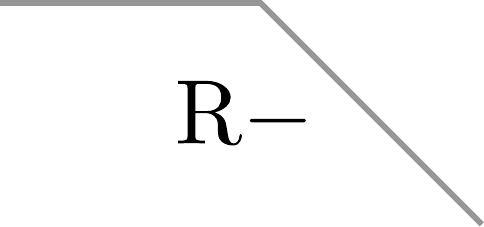}
        \caption{$-\relu(x)$}
    \end{subfigure}
    \vspace{-0.75em}
    \caption{The orientations of neurons captured by $L+$, $L-$, $R+$, and $R-$ for qualifying the scope of imprecision-contributions.}
    \vspace{-1em}
    \label{fig:quadrants}
\end{figure}

\begin{definition}
Given a single-layer \relu-network, $f$ with $n$ neurons and pre/post-activation weights $N,M$,
the {\em imprecision-contributions} of $f$ at $x$ are:
\begin{align*}
\sA{D}{S}{x} &\defeq \sum_{N_i x + b_i \geq 0 \wedge M_i \in S \wedge N_i \in D} |M_iN_i|, \\ \text{and }\;
\sP{D}{S}{x} &\defeq \sum_{N_i x + b_i \geq 0 \wedge M_i \in S \wedge N_i \in D} M_iN_i
\end{align*}
where $D$ can be the set $L \defeq \reals_{\leq 0}$, the set $R \defeq \reals_{\geq 0}$ or $\reals$
and $S$ can be the set $+ \defeq \reals_{\geq 0}$, the set $- \defeq \reals_{\leq 0}$ or $\reals$.
\end{definition}

Intuitively, $L+, L-, R+, R-$ correspond to the orientations that a neuron can take, as visualized in \cref{fig:quadrants}. 
$L$ (resp. $R$) results in contributions from neurons that activate as the argument $x$ of the imprecision-contribution function decreases (resp. increases).
Note that $f'(x) = \sP{\reals}{\reals}{x}$ if the derivative of $f$ is defined at $x$.

\begin{restatable}[End-Neuron Imprecision-Bound]{lemma}{AsymLemmaA} 
\label{lemma:asym_lb}
For all $\kappa \geq 0$ and single-layer \relu-networks, $f$, that classify $k$-flips 
for $k= \lceil \kappa \rceil + 5$, we have $\kappa < \max\{\sA{L}{\reals}{\flipX_1}, \sA{R}{\reals}{\flipX_k}\}$.
\end{restatable}
\begin{proofsketch}{AsymLemmaA}
We prove this by induction on $c \defeq \lfloor \kappa \rfloor$, using two simultaneous inductive invariants:
\begin{align*}
c & \leq \sA{L}{+}{\flipX_2} - \sA{R}{+}{\flipX_2} + \sA{R}{+}{\flipX_k} - \sA{L}{+}{\flipX_k}, \\ \text{and }\;
c & \leq \sA{L}{-}{\flipX_1} - \sA{R}{-}{\flipX_1} + \sA{R}{-}{\flipX_{k-1}} - \sA{L}{-}{\flipX_{k-1}}.
\end{align*}

This proof involves two key observations: 
(i) once imprecision-contribution in a direction has accumulated, it will only be larger for points further 
in that direction,
(ii) one must measure not just the accumulated growth of the imprecision-contribution at the ends of the approximated data ($\flipX_1$ and $\flipX_k$) in the out-wards directed neurons, but the growth of the {\em relative} imprecision-contribution excluding contribution from in-wards directed neurons.
We make these observations more precise in the full proof in the appendix.

\end{proofsketch}

Before demonstrating our main result, we require one further lemma, used to find a specific data-point with enough accumulated imprecision contribution to cause a violation:

\begin{restatable}[Lower-Bound on Imprecision-Contribution]{lemma}{AsymLemmaB}
\label{lemma:gen_lb}
For any $a\leq 1$ and $k \geq \lceil \frac{2}{a} \rceil + 5$
and single-layer \relu-network, $f$, that classifies $k$ flips,
there is some point $j \in [k]$ such that 
\[
\flipL_j a^{-1}( f(\flipX_j + a) + f(\flipX_j - a)) < \sA{\reals}{\reals}{\flipX_j} + \sA{R}{\reals}{\flipX_j - a} + \sA{L}{\reals}{\flipX_j + a}.
\]
\end{restatable}
\begin{proofsketch}{AsymLemmaB}
By using $c \defeq \frac{2}{a}$ we can apply \cref{lemma:asym_lb} (with $\kappa = c$), to show bounds for either the left or right-most \flipX{} (i.e., $j=1$ or $j=k$).  
For this point, we use the knowledge that the function is continuous, piecewise differentiable, to find points 
$l \in [\flipX_j - a, \flipX_j]$ and $u \in [\flipX_j, \flipX_j + a]$ such that $\frac{f(\flipX_j) - f(\flipX_j - a)}{a}< f'(l)$
and $f'(u) < \frac{f(\flipX_j + a) - f(\flipX_j)}{a}$ 
so we can use that
\[
f'(l) - f'(u) = \sP{\reals}{\reals}{l} - 
\sP{\reals}{\reals}{u} \leq \sA{L}{\reals}{\flipX_j - a} - 
\sA{R}{\reals}{\flipX_j + a},
\]
that $-c,c < \sA{\reals}{\reals}{\flipX_j},$ and that
\[
\flipL_j a^{-1}( f(\flipX_j + a) + f(\flipX_j - a)) 
= \frac{f(\flipX_j) - f(\flipX_j - a)}{a} - \frac{f(\flipX_j + a) - f(\flipX_j)}{a} - c
\] 
to produce the final upper bound.

\end{proofsketch}

We are now ready to show our main theorem, an upper bound on the number of flips that can be 
provably robustly classified with a single layer network.

\begin{restatable}[Single-Layer Limit]{theorem}{SingleLayerLimit}
\label{thm:ub} No single-layer \relu-network can provably $\alpha$-robustly 
classify $\lceil \frac{2}{\alpha}\rceil + 5$ or more flips for any $\alpha \in (0,1].$ 
\end{restatable}
\begin{proofsketch}{SingleLayerLimit}
The proof is by direct application of \cref{lemma:gen_lb}, and expansion of the definition of the derivative.  
We demonstrate that for the \flipX{} found by \cref{lemma:gen_lb}, the center of the box must be strictly closer to 0 than it's radius.
\end{proofsketch}

\section{Completely Interval Provable Classifiers are Impossible}
\label{sec:impossible}

Here we show our main result, that no neural network can be completely provably robust with interval 
analysis for simple functions.
We first introduce the necessary lemmas and machinery that allow us show a relationship 
between whether the network represents an invertible function, and where there is approximation error.

\begin{figure}[b]
\centering
\includegraphics[width=0.8\textwidth, trim=0cm 1cm 0cm 0cm]{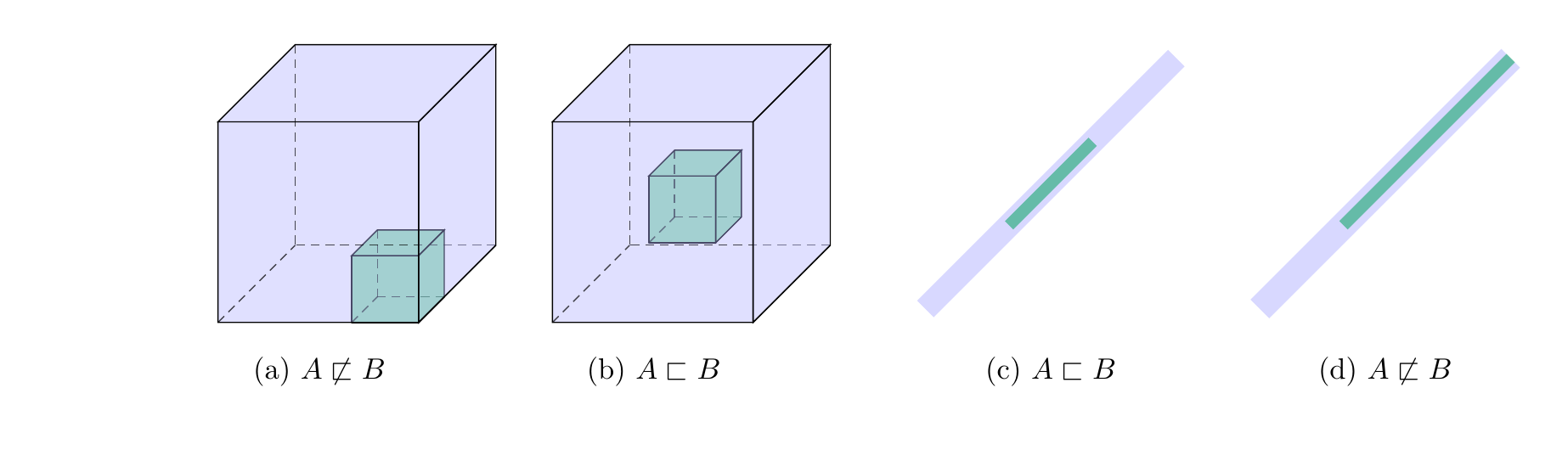}
\caption{Visualization of the relative interior relation.  The green (\greenbox) boxes are $A$ and the purple (\purplebox) boxes are $B$.}
\label{fig:relint}
\end{figure}

Counterintuitively, rather than being able to show that the transformed network in imprecise for a specific input box (i.e., that for a specific box $B$, we know $f[B] \subsetneq f^\#(B)$), we must, for any input box $B$ containing non-invertible points on its surface, {\em find} an input box, $A$, that is a strict subsets of $B$ (by a particular notion of strict defined below), such that $f(B) \subseteq f^\#(A)$.  
The fact that $A$ is a very strict subset of $B$ implies that interval analysis is imprecise enough on the network such that it can not be used to prove desired properties of $B$ (such that $f$ is completely robust for $B$).  
It is however crucial that $A$ not be required to be {\em too} strict a subset of $B$.  
One might be tempted to find subsets of the topological interior of $B$.  This however leads to significant technical issues: 
we need to have a notion of strict subset that applies even when some of the neurons in the network are unused (and zero). One can imagine the set representing the possible activations of those neurons as a lower dimensional surface embedded in a higher dimensional space, as in the case of \cref{fig:relint}(c) and (d). In this case, the interior of $B$ would be empty, even though we might have identified a subset of it that induces imprecision.

\subsection{The Relative Subset Relation}

We begin by formalizing the intuitive concept from \cref{fig:relint} using the notion of {\em relative interior}, and demonstrating some useful lemmas related to it.
First, recall for a set $S \subseteq \reals^d$ that the {\em affine hull} of $S$, written $\aff(S)$ is the smallest linear-subspace of $\reals^d$ that contains $S$. 

\begin{definition} We define {\em relative interior} as
$\relint(S) \defeq \{ x \in S \given \exists \epsilon >0 \st \hood_\epsilon(x) \cap \aff(S) \subseteq S \}
$
\end{definition}

We note that if $S \in \balls^d$, the set of closed, non-empty, axis-aligned boxes
of dimension $d$, we can restate the relative interior as 
$\relint(S) = \{ x \in S \given \forall i \in [d] \wehave x \in S|_i^\circ \cup \{\bcenter{S}_i\} \},$
where $S|_i^\circ$ is the interior of $S$'s restriction to dimension $i$.  

\begin{definition}[Relative Subset] $A$ is a {\em relative subset} of $B$, written $A \sqsubset B$,  
if and only if $A \subseteq \relint(B)$.
\end{definition}

We note again, that if $A,B \in \balls^d$, we can rephrase $A \sqsubset B$ as follows:  $A\subseteq B$ and
for each dimension, $i$, where $B|_i^\circ$ isn't empty,
$A|_i \subseteq B|_i^\circ$ or more concisely,
$\forall i \in [d] \wehave (B|_i^\circ \neq \emptyset \implies A|_i \subseteq B|_i^\circ).$
In particular, in one dimension, for real intervals $[a,b]$ and $[b,c]$ we have $[a,b] \sqsubset [a',b']$ if and only 
if $a < a' \leq b' < b$ or $a = a' = b' = b$.

Let $A,A',B,B',C$ be bounded and non-empty subsets of $\mathbb{R}^d$ in the following 
lemmas (the proofs of which can be found in \cref{appendix:relint}):

\begin{restatable}[Respects Projection]{lemma}{ProjLemma}
\label{lemma:rel:proj}
$A\sqsubset B$ implies $A|_i \sqsubset B|_i$.
\end{restatable}
\vspace{-1em}
\begin{restatable}[Respects Cartesian Product]{lemma}{CartesianLemma}
\label{lemma:rel:cartesian}
$A\sqsubset B$ and $A'\sqsubset B'$ implies $A\times A' \sqsubset B \times B'$.
\end{restatable}
\vspace{-1em}
\begin{restatable}[Downward Union]{lemma}{DownUnionLemma}
\label{lemma:rel:down_union}
$A\sqsubset C$ and $B\sqsubset C$ implies $A\cup B \sqsubset C$.
\end{restatable}
\vspace{-1em}
\begin{restatable}[Downward Hull]{lemma}{DownHullLemma}
\label{lemma:rel:down_hull}
$C \in \balls^d$ and $A\sqsubset C$ implies $\bHull(A) \sqsubset C$.
\end{restatable}

The following two trivial lemmas are trivial, and we frequently use them without mention:

\begin{lemma}[Singleton Reflexivity] %
\label{lemma:rel:sing}
$\{a\} \sqsubset \{a\}$.
\end{lemma}
\vspace{-1em}
\begin{lemma}[Center-Singleton is Always a Relative Subset] %
\label{lemma:rel:center}
$A \in \balls^d$ implies $\{\bcenter{A}\} \sqsubset A$.
\end{lemma}

It is important to note that some simple related properties counterintuitively do not always hold. Namely, 
if $A \sqsubset B$ and $B \subseteq C$ it is not always the case that $A \sqsubset C$.  
Furthermore, if $A \sqsubset B$ and $A' \sqsubset B'$ it is not always the case that $\bHull(A \cup A') \sqsubset \bHull(B \cup B')$.

\subsection{Inversion With Respect to The Relative Subset Relation}
Here we demonstrate that neural networks can loosely invert sets with respect to the relative subset relation. 
More formally, for any neural network $f$ with \relu-activations, one can essentially always find a strict 
subset, $X'$ of the relative interior of a box $X$ that the neural network maps to a superset 
of a specified subset $Y$ of the relative interior of the \boxhull{} of $f(X)$.  

\begin{restatable}[Concrete Relative Inversion]{lemma}{InversionLemma}
\label{lemma:inversion}
Suppose $f$ is a feed-forward network with \relu-activations and $Y,X' \in \balls^d$ 
and $X$ is compact and non-empty. Then

\centerline{
$Y \sqsubset \bHull(f(X)) 
\implies
\exists X' \sqsubset \bHull(X) \st Y \subseteq f^\#(X').$}
\end{restatable}
\begin{StrProofSketch}{InversionLemma}
The proof is by structural induction on the construction of $f$.  We use the lemma itself as the induction hypothesis.  Below we outline three key cases of the structural induction: sequential computations, relational parallel computations, and \relu{}:

\StrIndCaseSketch{Sequential Computation}{$f = g \circ h$.}
\begin{subproofsketch}
  By definition, $Y \sqsubset \bHull(g \circ h(X))$.
  Thus, there is some $H \sqsubset \bHull(h(X))$ 
  such that $Y \subseteq g^\#(H)$ by the induction hypothesis on $g$.  
  Applying the induction hypothesis again with the network $h$, 
  we get a set $X'\sqsubset \bHull(X)$ such that $H \subseteq h^\#(X')$.
  Thus, $Y \subseteq g^\#(H) \subseteq g^\#\circ h^\#(X') = f^\#(X')$.
\end{subproofsketch}

\StrIndCaseSketch{Relational Duplication}{$f(x) = (x, x)$.}
\begin{subproofsketch}
  In this case, we know $Y|_1 \sqsubset \bHull(X)$ and $Y|_2 \sqsubset \bHull(X)$ by
  \cref{lemma:rel:proj}.
  By \cref{lemma:rel:down_union}, we know $X'\defeq \bHull(Y|_1 \cup Y|_2)$ 
  is a relative subset of $\bHull(X)$, and also that
  $Y \subseteq X' \times X' = f^\#(X')$.
\end{subproofsketch}

\StrIndCaseSketch{Non-Relational Parallel Computation}{$f(x_1,x_2) = (g_1(x_1), g_2(x_2))$.}
\begin{subproofsketch}
      We know $Y|_1 \sqsubset \bHull(g_1(X|_1))$ and $Y|_2 \sqsubset \bHull(g_1(X|_1))$,
      so we can apply the induction hypothesis twice to produce $L \sqsubset \bHull(X|_1)$ 
        and $R \sqsubset \bHull(X|_2)$ such that 
        $Y|_1 \subseteq g^\#_1(L)$ and $Y|_2 \subseteq g^\#_1(R)$.
      We choose $X' = L \times R$. 
      By \cref{lemma:rel:cartesian},
      we have $X' \sqsubset \bHull(X)$.
      Then $Y \subseteq g^\#_1(X'|_1) \times g^\#_2(X'|_2) = f^\#(X')$.
    \end{subproofsketch}

\StrIndCaseSketch{Activation}{$f(x) = \relu(x)$.}
\begin{subproofsketch}
  We know here that $f \from \reals \to \reals$ which simplifies the proof. 
  In this case, \boxhull{} of $f(X)$ must either be a subset of $\reals_{\geq 0}$ so either $Y$ is the singleton set containing zero, or a subset of $\reals_{> 0}$.  In the first case, we pick $X'$ to be an easy-to-pick (the singleton set containing center as in \cref{lemma:rel:center}) relative subset of the hull of $X$.  
Otherwise, we can pick $Y$ itself, since $\relu(Y) = Y \subseteq \bHull(X)$.
\end{subproofsketch}
\end{StrProofSketch}

\subsection{Impossibility for Non-Invertibility}
\label{subsec:impossible}

\DeclareRobustCommand\NonInvBox{\tikz{\draw[fill=\NonInvColor, opacity=1, draw=none] (6.69,0) rectangle ++(0.15,0.15);}}
\DeclareRobustCommand\HNBox{\tikz{\draw[fill=\HNColor, opacity=0.8, draw=none] (6.69,0) rectangle ++(0.15,0.15);}}
\DeclareRobustCommand\StdBox{\tikz{\draw[fill=\StdColor, opacity=1, draw=none] (6.69,0) rectangle ++(0.15,0.15);}}
\DeclareRobustCommand\FHNBox{\tikz{\draw[fill=\FHNColor, opacity=1, draw=none] (6.69,0) rectangle ++(0.15,0.15);}}
\DeclareRobustCommand\FMBox{\tikz{\draw[fill=\FMColor, opacity=0.5, draw=none] (6.69,0) rectangle ++(0.15,0.15);}}

\begin{figure}[b]
\centering
\begin{minipage}[c]{0.48\textwidth}
\caption{A visualization of the claims of \cref{theorem:imprecision}. 
The neural network, $f$ is classifying three flips  $\flipX_{-1}, \flipX_{0}$ and $\flipX_1$ (i.e., $D = \{(-2,-1), (0,1), (2,-1)\}$ as in \cref{thm:ex}).
Here, $x=0$ is the non-invertability and $f^{-1}(x)=\{-1,1\}$. These two points mapping to $x$, which constitute $N \subseteq f^{-1}(x)$, are marked in red (\NonInvBox). The \boxhull, $\bHull(N)$, is marked in green (\HNBox).  
The perfect approximation of $f$ on $\bHull(N)$ (i.e., $\{ (v,f(v)) \given v \in N\}$)
is marked in blue (\FHNBox).   
We can see that the interval approximation of the relative interior box $M \sqsubset \bHull(N)$ looses precision by looking at the orange region (\FMBox).
}
\label{fig:main_thm}
\end{minipage}
\hfill
\begin{minipage}[c]{0.51\textwidth}
\includegraphics[width=\textwidth, trim=0.4cm 0.2cm 0.5cm 0cm]{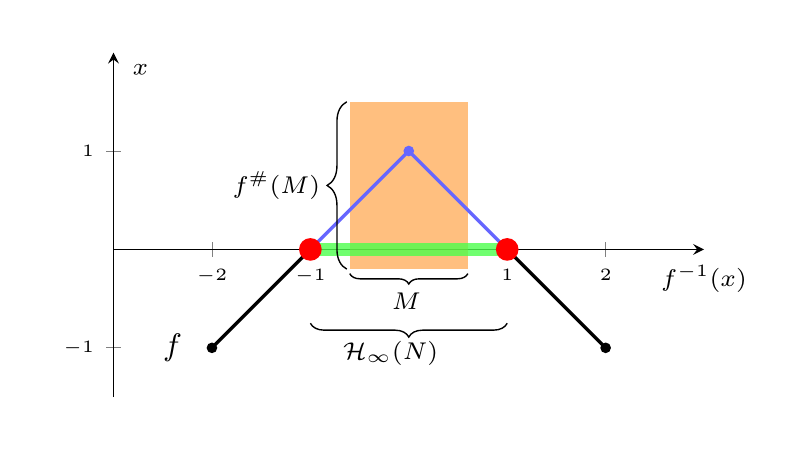}
\end{minipage}
\end{figure}

We now prove our central result, that non-invertible neural networks necessarily induce approximation imprecision.
Essentially, as visualized in \cref{fig:main_thm}, we show that there is a box, $M$ which is a relative subset (i.e., usually very strict subset) 
of the \boxhull{} of any region for which the network is entirely not injective, 
such that analyzing the network with $M$ includes any of the non-invertible 
points in the inferred approximation.

The key idea is that while there might be non-invertible points on the boundary of a region, 
$M$ does not include those points, since it is a relative subset. 
Thus, we can use this theorem to infer areas where analyzing the network produces approximations that include points
that aren't in the concrete, or true, set of possible network outputs. 

\begin{restatable}[Non-Invertibility Induces Interval Imprecision]{theorem}{ImprecisionThm}
\label{theorem:imprecision}
Suppose $f \from \mathbb{R}^n \to \mathbb{R}^m$ is a feed-forward network with \relu-activations and 
$x \in \mathbb{R}^m$ and $N \subseteq f^{-1}(x)$ is compact and non-empty.  Then (assuming $M$ is a box):

\centerline{$\exists M \sqsubset \bHull(N) \ldot x \in f^\#(M).$}
\end{restatable}
\begin{StrProofSketch}{ImprecisionThm}
Again, the proof is by structural induction on the construction of $f$, using the lemma itself as the induction hypothesis.  Below we outline three key cases: sequential computations, relational parallel computations, and addition:

\StrIndCaseSketch{Sequential Computation}{$f = g \circ h$}
\begin{subproofsketch}
  We know that $N \subseteq h^{-1} \circ g^{-1}(x)$, and thus
  can use the induction hypothesis on $h$ to produce $M' \sqsubset \bHull(h(N))$ such that $x \in g^\#(M')$.
  By \cref{lemma:inversion}, we know that there is some $M\sqsubset \bHull(N)$ such that $M' \subseteq h^\#(M)$ and 
  thus that $x \in g^\# \circ h^\#(M)$.
\end{subproofsketch}

\StrIndCaseSketch{Relational Duplication}{$f(y) = (y, y)$.}
\begin{subproofsketch}
  We have $N\subseteq \{x_1\} \cap \{x_2\}$ so $N=\{x_1\}$. By singleton reflexivity, $N \sqsubset \bHull(N)$.  
  Thus, $x \in f^\#(N)$.
\end{subproofsketch}

\StrIndCaseSketch{Non-Relational Parallel Computation}{$f(x_1,x_2) = (g_1(x_1), g_2(x_2))$.}
\begin{subproofsketch}
      First we know $N|_1 \subseteq g_1^{-1}(x_1)$ and $N|_2 \subseteq g_2^{-1}(x_2)$ 
      by projection and that $N|_1$ and $N|_2$ are still compact and non-empty.
      Thus, by the induction hypothesis twice we see that there are boxes
      $M_1 \sqsubset \bHull(N|_1)$ and $M_2 \sqsubset \bHull(N|_2)$ 
      such that 
      $x_1 \in g_1^\#(M_1)$ and
      $x_2 \in g_2^\#(M_2)$.
      Then $M_1 \times M_2 \sqsubset \bHull(N)$ by 
        \cref{lemma:rel:cartesian}.
      Then $x_1 \in g_1^\#(M_1 \times M_2)$ 
      and $x_2 \in g_2^\#(M_1 \times M_2)$ 
      by soundness.
      Thus, there is some box $M\sqsubset \bHull(N)$ such that $x \in f^\#(M)$.
\end{subproofsketch}

\StrIndCaseSketch{Addition}{$f(y_1,y_2) = y_1 + y_2$.}
\begin{subproofsketch}
  Because $f \from \reals^2 \to \reals$, we know $f^{-1}(x) = \{ (a, x - a) \given a \in \mathbb{R}\}$. 
  We can pick $M = \{ \bcenter{\bHull(N)} \}$ and demonstrate that $M \sqsubset \bHull(N)$ and that $x \in f^\#(M)$.
\end{subproofsketch}

\end{StrProofSketch}

\subsection{Implications and Corollaries}
\label{here}

Here we demonstrate implications of this result, such as that no neural network can completely (as defined below), 
and provably robustly with interval, classify every dataset.

\begin{definition}
\label{def:complete}
We say that a neural network $f \from \mathbb{R}^d \to \mathbb{R}$ is a 
{\em completely $\nu$-(provable) robust classifier} for points $x_1, \ldots, x_n \in \mathbb{R}^d$ 
and labels $l_1,\ldots, l_n \in \{-1,1\}$
if given $\delta = \frac{1}{2}\min_{i \neq j} ||x_i - x_j||_\infty$ then 
$\forall \epsilon < \nu\delta \wehave \forall i \in [d] \wehave f(\ball{\epsilon}{x_i}) l_i > 0.$
If it is {\em provable} then we also have that $\forall i \in [d] \wehave f^\#(\ball{\epsilon}{x_i}) l_i > 0.$
\end{definition}

We note that the specification task implicitly induced by this definition is complete as defined by the more general notion in the introduction.

\begin{restatable}[Completely Provable Classifiers are Impossible]{corollary}{CompleteTheorem}
There is no feed forward \relu-network that is a completely $1$-provable classifier for the 
dataset $D = \{(-2,-1), (0,1), (2,-1)\}$.
\label{thm:ex}
\end{restatable}
\begin{iproof}
Suppose $f \from \mathbb{R} \to \mathbb{R}$ is a completely $1$-provable 
interpolator for this dataset. Then by continuity we know that $f(-1) = f(1) = 0$,
and thus that $\{-1,1\} \subseteq f^{-1}(0)$ (which is a compact non-empty set).
Then by application of \cref{theorem:imprecision}, 
there is some set $M \sqsubset \bHull(\{ -1,1\})$ such that $0 \in f^\#(M)$.
Rephrased, this means that there are $a,b \in (-1,1)$ such that $0 \in f^\#([a,b])$.
This contradicts with the definition of a completely $1$-provable classifier however.
\end{iproof}

\begin{proposition}[Single-Layer Completely Robust Classifiers Always Exist]
For {\em any} dataset of $n$ points $x_i \in \reals$ and labels $l_i \in \{-1,1\}$ 
there is a one-hidden-layer \relu-network that completely robustly (but not necessarily provably) classifies it.
\label{prop:existance}
\end{proposition}
\begin{iproof}
We present the construction explicitly. Let $\delta \defeq \min\{ |x_i-x_j| \given i\neq j\}$ in
\vspace{-0.5em}
\[
f(y) \defeq \sum_{i=1}^{n} l_{i}\left( 
  \relu\left[ \frac{1}{\delta}(y - (x_i - \delta)) \right]
- \relu\left[ \frac{2}{\delta}(y - x_i) \right]
+ \relu\left[ \frac{1}{\delta}(y - (x_i + \delta)) \right] 
   \right)
\]
One can check that this works by plugging in $x_j$, although a full proof is by induction.  
While this is not immediately of the form described for one-hidden layer networks, 
one can see easily how to algebraically convert this into that form.  Because we only care about the robustness, 
and not interval provability of $f$, this is sufficient.
\end{iproof}

\section{Discussion and Future Work}

While we limited the scope of our discussion to \relu-activations, 
we note that our theorems extetend trivially to any monotone bounded activation.  
However, we observe that non-monotonic activations functions (such as absolute value) 
do not admit the same forms of theorems.  
Our preliminary experiments however indicate that substituting \relu{} with these 
activation functions does not result in easier training or provability.
This suggests that there are more general version of the the theorems presented here, 
in particular relating the difficulty of program synthesis with the relational expressiveness 
of the domain used to verify the specification.

\section{Conclusion}

In this paper, we proved two theorems that show limits in the expressiveness of interval
provable neural networks. 
We showed that no \relu-network can completely provably classify simple
one-dimensional datasets containing only three points. This indicates a
fundamental loss of precision whenever \relu-networks are
analyzed using interval arithmetic, which can not be regained, no matter the network. 
Further, we showed that a single hidden layer \relu-network can not provably
classify simple datasets even without the requirement for completeness, 
which is in stark contrast to classical universal approximation theorems, 
where a single hidden layer is sufficient. This shows
that the approximative capabilities of interval provable networks are lower
compared to standard neural networks.

\begin{acks}
We thank Joseph Swernofsky for his helpful comments on earlier drafts of this work.
\end{acks}

\bibliographystyle{alpha} %
\bibliography{ms}

\newcommand{\etalchar}[1]{$^{#1}$}
\begin{thebibliography}{WPW{\etalchar{+}}18}

\bibitem[Ano22]{anonymous2022on}
Anonymous.
\newblock On the convergence of certified robust training with interval bound
  propagation.
\newblock In {\em Submitted to The Tenth International Conference on Learning
  Representations}, 2022.
\newblock under review.

\bibitem[BBS{\etalchar{+}}19]{Balunovic2019}
Mislav Balunovic, Maximilian Baader, Gagandeep Singh, Timon Gehr, and Martin
  Vechev.
\newblock Certifying geometric robustness of neural networks.
\newblock In {\em NeurIPS}, 2019.

\bibitem[BIL{\etalchar{+}}16]{BastaniILVNC16}
Osbert Bastani, Yani Ioannou, Leonidas Lampropoulos, Dimitrios Vytiniotis,
  Aditya~V. Nori, and Antonio Criminisi.
\newblock Measuring neural net robustness with constraints.
\newblock In {\em NeurIPS}, 2016.

\bibitem[BMV20]{baader2020universal}
Maximilian Baader, Matthew Mirman, and Martin Vechev.
\newblock Universal approximation with certified networks.
\newblock {\em ICLR}, 2020.

\bibitem[BTD{\etalchar{+}}16]{BojarskiTDFFGJM16}
Mariusz Bojarski, Davide~Del Testa, Daniel Dworakowski, Bernhard Firner, Beat
  Flepp, Prasoon Goyal, Lawrence~D. Jackel, Mathew Monfort, Urs Muller, Jiakai
  Zhang, Xin Zhang, Jake Zhao, and Karol Zieba.
\newblock End to end learning for self-driving cars.
\newblock {\em arXiv preprint arxiv:1604.07316}, 2016.

\bibitem[BV20]{balunovic2019adversarial}
Mislav Balunovic and Martin Vechev.
\newblock Adversarial training and provable defenses: Bridging the gap.
\newblock In {\em ICLR}, 2020.

\bibitem[BWC{\etalchar{+}}19]{Boopathy2019}
Akhilan Boopathy, Tsui-Wei Weng, Pin-Yu Chen, Sijia Liu, and Luca Daniel.
\newblock Cnn-cert: An efficient framework for certifying robustness of
  convolutional neural networks.
\newblock In {\em AAAI}, volume~33, 2019.

\bibitem[BWL{\etalchar{+}}21]{boopathy2021fast}
Akhilan Boopathy, Tsui-Wei Weng, Sijia Liu, Pin-Yu Chen, Gaoyuan Zhang, and
  Luca Daniel.
\newblock Fast training of provably robust neural networks by singleprop.
\newblock {\em arXiv preprint arXiv:2102.01208}, 2021.

\bibitem[CAH18]{Croce2018}
Francesco Croce, Maksym Andriushchenko, and Matthias Hein.
\newblock Provable robustness of relu networks via maximization of linear
  regions.
\newblock {\em arXiv preprint arXiv:1810.07481}, 2018.

\bibitem[CC77]{CC77}
Patrick Cousot and Radhia Cousot.
\newblock Abstract interpretation: a unified lattice model for static analysis
  of programs by construction or approximation of fixpoints.
\newblock In {\em Symposium on Principles of Programming Languages (POPL)},
  1977.

\bibitem[CH19a]{croce2019provable}
Francesco Croce and Matthias Hein.
\newblock Provable robustness against all adversarial $l_p$-perturbations for
  $p >= 1$.
\newblock In {\em ICLR}, 2019.

\bibitem[CH19b]{croce2019sparse}
Francesco Croce and Matthias Hein.
\newblock Sparse and imperceivable adversarial attacks.
\newblock In {\em Proceedings of the IEEE/CVF International Conference on
  Computer Vision (ICCV)}, 2019.

\bibitem[CH20]{croce2020reliable}
Francesco Croce and Matthias Hein.
\newblock Reliable evaluation of adversarial robustness with an ensemble of
  diverse parameter-free attacks.
\newblock In {\em ICML}, 2020.

\bibitem[CW17]{Carlini017}
Nicholas Carlini and David~A. Wagner.
\newblock Towards evaluating the robustness of neural networks.
\newblock In {\em Symposium on Security and Privacy (SP)}, 2017.

\bibitem[Cyb89]{cybenko1989approximation}
George Cybenko.
\newblock Approximation by superpositions of a sigmoidal function.
\newblock {\em Mathematics of Control, Signals and Systems (MCSS)}, 1989.

\bibitem[GDS{\etalchar{+}}18]{Gowal2018}
Sven Gowal, Krishnamurthy Dvijotham, Robert Stanforth, Rudy Bunel, Chongli Qin,
  Jonathan Uesato, Timothy Mann, and Pushmeet Kohli.
\newblock On the effectiveness of interval bound propagation for training
  verifiably robust models.
\newblock {\em arXiv preprint arXiv:1810.12715}, 2018.

\bibitem[GMT{\etalchar{+}}18]{ai2}
Timon Gehr, Matthew Mirman, Petar Tsankov, Dana Drachsler~Cohen, Martin Vechev,
  and Swarat Chaudhuri.
\newblock Ai2: Safety and robustness certification of neural networks with
  abstract interpretation.
\newblock In {\em Symposium on Security and Privacy (SP)}, 2018.

\bibitem[GSS15]{Goodfellow2015}
Ian~J Goodfellow, Jonathon Shlens, and Christian Szegedy.
\newblock Explaining and harnessing adversarial examples.
\newblock In {\em ICLR}, 2015.

\bibitem[HJVE01]{hickey2001interval}
Timothy Hickey, Qun Ju, and Maarten~H Van~Emden.
\newblock Interval arithmetic: From principles to implementation.
\newblock {\em Journal of the ACM (JACM)}, 48(5):1038--1068, 2001.

\bibitem[HKR{\etalchar{+}}20]{huang2020survey}
Xiaowei Huang, Daniel Kroening, Wenjie Ruan, James Sharp, Youcheng Sun, Emese
  Thamo, Min Wu, and Xinping Yi.
\newblock A survey of safety and trustworthiness of deep neural networks:
  Verification, testing, adversarial attack and defence, and interpretability.
\newblock {\em Computer Science Review}, 37:100270, 2020.

\bibitem[HSW89]{hornik1989multilayer}
Kurt Hornik, Maxwell~B. Stinchcombe, and Halbert White.
\newblock Multilayer feedforward networks are universal approximators.
\newblock {\em Neural Networks}, 1989.

\bibitem[KBD{\etalchar{+}}17]{katz2017reluplex}
Guy Katz, Clark Barrett, David~L Dill, Kyle Julian, and Mykel~J Kochenderfer.
\newblock Reluplex: An efficient smt solver for verifying deep neural networks.
\newblock In {\em International Conference on Computer Aided Verification
  (CAV)}, 2017.

\bibitem[KGB16]{KurakinGB16}
Alexey Kurakin, Ian~J. Goodfellow, and Samy Bengio.
\newblock Adversarial examples in the physical world.
\newblock {\em arXiv preprint arxiv:1607.02533}, 2016.

\bibitem[LAL{\etalchar{+}}19]{liu2019algorithms}
Changliu Liu, Tomer Arnon, Christopher Lazarus, Christopher Strong, Clark
  Barrett, and Mykel~J Kochenderfer.
\newblock Algorithms for verifying deep neural networks.
\newblock {\em arXiv preprint arXiv:1903.06758}, 2019.

\bibitem[LSR{\etalchar{+}}21]{lin2020certified}
Wan-Yi Lin, Fatemeh Sheikholeslami, Leslie Rice, J~Zico Kolter, et~al.
\newblock Certified robustness against physically-realizable patch attack via
  randomized cropping.
\newblock In {\em ICLR}, 2021.

\bibitem[LSS21]{liu2021training}
Chen Liu, Mathieu Salzmann, and Sabine S{\"u}sstrunk.
\newblock Training provably robust models by polyhedral envelope
  regularization.
\newblock {\em IEEE Transactions on Neural Networks and Learning Systems},
  2021.

\bibitem[LTC19]{Liu2019}
Chen Liu, Ryota Tomioka, and Volkan Cevher.
\newblock On certifying non-uniform bound against adversarial attacks.
\newblock In {\em ICML}, 2019.

\bibitem[MGV18]{diffai}
Matthew Mirman, Timon Gehr, and Martin Vechev.
\newblock Differentiable abstract interpretation for provably robust neural
  networks.
\newblock In {\em ICML}, 2018.

\bibitem[Moo66]{moore1996interval}
Ramon~E Moore.
\newblock {\em Interval analysis}.
\newblock Prentice-Hall Englewood Cliffs, NJ, 1966.

\bibitem[PMG16]{papernot2016transferability}
Nicolas Papernot, Patrick McDaniel, and Ian Goodfellow.
\newblock Transferability in machine learning: from phenomena to black-box
  attacks using adversarial samples.
\newblock {\em arXiv preprint arXiv:1605.07277}, 2016.

\bibitem[PMW{\etalchar{+}}16]{PapernotMWJS15}
Nicolas Papernot, Patrick~D. McDaniel, Xi~Wu, Somesh Jha, and Ananthram Swami.
\newblock Distillation as a defense to adversarial perturbations against deep
  neural networks.
\newblock In {\em {IEEE} Symposium on Security and Privacy ({SP})}, 2016.

\bibitem[RSL18]{Raghunathan2018}
Aditi Raghunathan, Jacob Steinhardt, and Percy Liang.
\newblock Certified defenses against adversarial examples.
\newblock In {\em ICLR}, 2018.

\bibitem[SGM{\etalchar{+}}18]{Singh2018}
Gagandeep Singh, Timon Gehr, Matthew Mirman, Markus P{\"u}schel, and Martin
  Vechev.
\newblock Fast and effective robustness certification.
\newblock In {\em NeurIPS}, 2018.

\bibitem[SGPV19]{Singh2019}
Gagandeep Singh, Timon Gehr, Markus P{\"u}schel, and Martin Vechev.
\newblock An abstract domain for certifying neural networks.
\newblock In {\em POPL}, 2019.

\bibitem[SHS20]{stutz2020confidence}
David Stutz, Matthias Hein, and Bernt Schiele.
\newblock Confidence-calibrated adversarial training: Generalizing to unseen
  attacks.
\newblock In {\em ICML}, 2020.

\bibitem[SWZ{\etalchar{+}}21]{shi2021fast}
Zhouxing Shi, Yihan Wang, Huan Zhang, Jinfeng Yi, and Cho-Jui Hsieh.
\newblock Fast certified robust training with short warmup.
\newblock {\em Advances in Neural Information Processing Systems}, 2021.

\bibitem[SYN15]{ShahamYN15}
Uri Shaham, Yutaro Yamada, and Sahand Negahban.
\newblock Understanding adversarial training: Increasing local stability of
  neural nets through robust optimization.
\newblock {\em arXiv preprint arxiv:1511.05432}, 2015.

\bibitem[SYZ{\etalchar{+}}19]{Salman2019}
Hadi Salman, Greg Yang, Huan Zhang, Cho-Jui Hsieh, and Pengchuan Zhang.
\newblock A convex relaxation barrier to tight robustness verification of
  neural networks.
\newblock In {\em NeurIPS}, 2019.

\bibitem[SZS{\etalchar{+}}13]{adversarialDiscovery}
Christian Szegedy, Wojciech Zaremba, Ilya Sutskever, Joan Bruna, Dumitru Erhan,
  Ian~J. Goodfellow, and Rob Fergus.
\newblock Intriguing properties of neural networks.
\newblock {\em arXiv preprint arXiv:1312.6199}, 2013.

\bibitem[TKP{\etalchar{+}}17]{tramer2017ensemble}
Florian Tram{\`e}r, Alexey Kurakin, Nicolas Papernot, Ian Goodfellow, Dan
  Boneh, and Patrick McDaniel.
\newblock Ensemble adversarial training: Attacks and defenses.
\newblock {\em arXiv preprint arXiv:1705.07204}, 2017.

\bibitem[Tuc02]{tucker2002rigorous}
Warwick Tucker.
\newblock A rigorous ode solver and smale’s 14th problem.
\newblock {\em Foundations of Computational Mathematics}, 2(1):53--117, 2002.

\bibitem[WAPJ20]{wang2020interval}
Zi~Wang, Aws Albarghouthi, Gautam Prakriya, and Somesh Jha.
\newblock Interval universal approximation for neural networks.
\newblock In {\em POPL}, 2020.

\bibitem[WK18]{kolter2018provable}
Eric Wong and Zico Kolter.
\newblock Provable defenses against adversarial examples via the convex outer
  adversarial polytope.
\newblock 2018.

\bibitem[WPW{\etalchar{+}}18]{Wang2018a}
Shiqi Wang, Kexin Pei, Justin Whitehouse, Junfeng Yang, and Suman Jana.
\newblock Efficient formal safety analysis of neural networks.
\newblock In {\em NeurIPS}, 2018.

\bibitem[WRK19]{wong2019fast}
Eric Wong, Leslie Rice, and J~Zico Kolter.
\newblock Fast is better than free: Revisiting adversarial training.
\newblock In {\em ICLR}, 2019.

\bibitem[WSK19]{wong2019wasserstein}
Eric Wong, Frank Schmidt, and Zico Kolter.
\newblock Wasserstein adversarial examples via projected sinkhorn iterations.
\newblock In {\em ICML}, 2019.

\bibitem[WSMK18]{Wong2018}
Eric Wong, Frank Schmidt, Jan~Hendrik Metzen, and J~Zico Kolter.
\newblock Scaling provable adversarial defenses.
\newblock In {\em NeurIPS}, 2018.

\bibitem[XTSM19]{xiao2019training}
Kai Xiao, Vincent Tjeng, Nur~Muhammad Shafiullah, and Aleksander Madry.
\newblock Training for faster adversarial robustness verification via inducing
  relu stability.
\newblock In {\em International Conference on Learning Representations}, 2019.

\bibitem[XZW{\etalchar{+}}21]{xu2021fast}
Kaidi Xu, Huan Zhang, Shiqi Wang, Yihan Wang, Suman Jana, Xue Lin, and Cho-Jui
  Hsieh.
\newblock {Fast and Complete}: Enabling complete neural network verification
  with rapid and massively parallel incomplete verifiers.
\newblock In {\em International Conference on Learning Representations}, 2021.

\bibitem[ZAD21]{zhang2021certified}
Yuhao Zhang, Aws Albarghouthi, and Loris D'Antoni.
\newblock Certified robustness to programmable transformations in lstms.
\newblock In {\em EMNLP}, 2021.

\bibitem[ZCX{\etalchar{+}}20]{zhang2020towards}
Huan Zhang, Hongge Chen, Chaowei Xiao, Sven Gowal, Robert Stanforth, Bo~Li,
  Duane Boning, and Cho-Jui Hsieh.
\newblock Towards stable and efficient training of verifiably robust neural
  networks.
\newblock In {\em ICLR}, 2020.

\bibitem[ZWC{\etalchar{+}}18]{zhang2018crown}
Huan Zhang, Tsui-Wei Weng, Pin-Yu Chen, Cho-Jui Hsieh, and Luca Daniel.
\newblock Efficient neural network robustness certification with general
  activation functions.
\newblock In {\em Advances in Neural Information Processing Systems (NuerIPS)},
  dec 2018.

\end{thebibliography}

\appendix

\newpage
\section{Extended Proofs for Single Hidden Layer Network Results}
\label{appendix:single}
Here we restate the theorems and show the full proofs for the results in \cref{sec:single}.
\AsymLemmaA* 
\begin{iproof_of}{lemma:asym_lb}
\label{prf:AsymLemmaA}
We prove this by induction on $c$.

\prfcase{Induction Hypothesis}{ 
Given $c \in \nats \cup \{0,-1,-2\}$ 
there is some even natural number $k\leq \relu(c+2) + 2$ 
such that for any single-layer \relu-network, $f$ that classifies $k$ flips we have
\begin{align*}
c & \leq \sA{L}{+}{\flipX_2} - \sA{R}{+}{\flipX_2} + \sA{R}{+}{\flipX_k} - \sA{L}{+}{\flipX_k}, \\ \text{and }\;
c & \leq \sA{L}{-}{\flipX_1} - \sA{R}{-}{\flipX_1} + \sA{R}{-}{\flipX_{k-1}} - \sA{L}{-}{\flipX_{k-1}}.
\end{align*}
}

\prfcase{Base Case}{Suppose $c \leq 0$.}
\begin{subproof}
  Pick $k=2$. Then 
\begin{align*}
\sA{L}{+}{\flipX_2} - \sA{R}{+}{\flipX_2} + \sA{R}{+}{\flipX_2} - \sA{L}{+}{\flipX_2} &= 0 \geq c, \;\text{ and } \\ 
\sA{L}{-}{\flipX_1} - \sA{R}{-}{\flipX_1} + \sA{R}{-}{\flipX_{2 - 1}} - \sA{L}{-}{\flipX_{2 - 1}} &= 0 \geq c.
\end{align*}
\vspace{-3.75em}

\end{subproof}

\prfcase{Induction Step}{Suppose $c > 0$, and the induction hypothesis holds for $c - 2$.}
\begin{subproof}
Then there is some even natural $k' \leq \relu(c - 2 + 2) + 2$ 
such that for any single-layer \relu-network, $f$, that is a classifier for $k'$ flips we have
\begin{align*}
c-2 & \leq \sA{L}{+}{\flipX_2} - \sA{R}{+}{\flipX_2} + \sA{R}{+}{\flipX_{k'}} - \sA{L}{+}{\flipX_{k'}}, \\ \text{and }\;
c-2 & \leq \sA{L}{-}{\flipX_1} - \sA{R}{-}{\flipX_1} + \sA{R}{-}{\flipX_{k'-1}} - \sA{L}{-}{\flipX_{k'-1}}.
\end{align*}

Pick $k = k' + 2$.  Then $k$ is even, and $k \leq \relu(c) + 4 \leq \relu(c + 2) + 2$ since $c > 0$.

Let $f$ be any single-layer \relu-network that classifies $k$ flips.  Then $f$ also classifies $k'$ flips.

We only show the positive bound, that $c \leq \sA{L}{+}{\flipX_2} - \sA{R}{+}{\flipX_2} + \sA{R}{+}{\flipX_k} - \sA{L}{+}{\flipX_k}$.

The proof for the negative bound is analogous. 

There must be some point $l \in [\flipX_{k'},\flipX_{k'+1}]$ such that $f'(l) \leq -1$ by the mean value theorem 
(since \relu-networks are continuous) and because $f(\flipX_{k'}) = 1 = -f(\flipX_{k'+1})$.

Similarly, there must be some point $u \in [\flipX_{k'+1}, \flipX_k]$ such that $1 \leq f'(u)$. Thus, 
\begin{align*}
2 & \leq f'(u) - f'(l) \\
  & \leq \sP{L}{+}{u} + \sP{L}{-}{u} + \sP{R}{+}{u} + \sP{R}{-}{u} \\
  &    - \sP{L}{+}{l} - \sP{L}{-}{l} - \sP{R}{+}{l} - \sP{R}{-}{l}.
\end{align*}

We know $ \sP{L}{-}{u} -\sP{L}{-}{l} \leq 0$ and $\sP{R}{-}{u} - \sP{R}{-}{l} \leq 0$ 
and $\sP{L}{+}{t} = -\sA{L}{+}{t}$ and $\sP{R}{+}{t} = \sA{R}{+}{t}$ for any $t$ so 
\begin{align*}
2 & \leq -\sA{L}{+}{u} + \sA{R}{+}{u}\\
  &      + \sA{L}{+}{l} - \sA{R}{+}{l}.
\end{align*}

We also know $\sA{L}{S}{t}$ increases as $t$ decreases and $\sA{R}{S}{t}$ increases as $t$ increases, so
\begin{align*}
2 & \leq -\sA{L}{+}{\flipX_k} + \sA{R}{+}{\flipX_k}\\
  &     + \sA{L}{+}{\flipX_{k'}} - \sA{R}{+}{\flipX_{k'}}.
\end{align*}

By combining with the positive inductive bound, we get
\begin{align*}
(c-2)+2 = c & \leq \sA{L}{+}{\flipX_2} - \sA{R}{+}{\flipX_2} + \sA{R}{+}{\flipX_{k'}} - \sA{L}{+}{\flipX_{k'}} \\
            &      -\sA{L}{+}{\flipX_k} + \sA{R}{+}{\flipX_k} + \sA{L}{+}{\flipX_{k'}} - \sA{R}{+}{\flipX_{k'}} \\
            & \leq \sA{L}{+}{\flipX_2} - \sA{R}{+}{\flipX_2} -\sA{L}{+}{\flipX_k} + \sA{R}{+}{\flipX_k},
\end{align*}

which proves the positive bound of the induction hypothesis for $c$.
\end{subproof}

Thus, by induction, we find that there is some $k \leq \lceil c \rceil + 5$ such that, 
after removing the negative terms and increasing by swapping $\flipX_2$ with $\flipX_1$ and $\flipX_{k-1}$ with $\flipX_k$:
\begin{align*}
c + 1 & \leq \sA{L}{+}{\flipX_1} + \sA{R}{+}{\flipX_k}, \\ \text{and }\;
c + 1 & \leq \sA{L}{-}{\flipX_1} + \sA{R}{-}{\flipX_k}.
\end{align*}

Summing these equations together gives us:
\[
2c + 2 \leq \sA{L}{+}{\flipX_1} + \sA{R}{+}{\flipX_k} + \sA{L}{-}{\flipX_1} + \sA{R}{-}{\flipX_k} = 2 \max \{ \sA{L}{\reals}{\flipX_1}, \sA{R}{\reals}{\flipX_k} \},
\]

and thus that $c < \max \{ \sA{L}{\reals}{\flipX_1}, \sA{R}{\reals}{\flipX_k}.$
\end{iproof_of}

\AsymLemmaB*
\begin{iproof_of}{lemma:gen_lb}
\label{prf:AsymLemmaB}
Let $k \geq \lceil 2a^{-1} \rceil + 5, $ and define $c \defeq \frac{2}{a}.$
Because $k \geq \lceil |c| \rceil + 5$, we can use \cref{lemma:asym_lb} 
to get that $|c| < \max \{ \sA{L}{\reals}{\flipX_1}, \sA{R}{\reals}{\flipX_k} \}$.

For convenience, we define $\tilde{f}_L(x) = a^{-1}(f(x) - f(x-a))$ and $\tilde{f}_R(x) = a^{-1}(f(x + a) - f(x))$.

We only show the proof when $|c| < \sA{L}{\reals}{\flipX_1}$, the other case is analogous, but picking $j = k$.

In this case we know $0 < \sA{L}{\reals}{\flipX_1}$ and thus, 
\[
\flipL_1 a^{-1}( f(\flipX_1 + a) + f(\flipX_1 - a)) \leq \tilde{f}_L(\flipX_1) - \tilde{f}_R(\flipX_1) - c.
\]

There must be a point $l \in [\flipX_1 - a, \flipX_1]$ such that $\tilde{f}_L(l) \leq f'(l)$ 
and a point $u \in [\flipX_1, \flipX_1 + a]$ such that $f'(u) \leq \tilde{f}_R(u).$

We can thus derive, in a manner similar to what is seen in \cref{lemma:asym_lb}:
\begin{align*}
\tilde{f}_L(\flipX_1) - \tilde{f}_R(\flipX_1) 
  \leq f'(l) - f'(u)
 & \leq \sP{L}{+}{l} + \sP{L}{-}{l} + \sP{R}{+}{l} + \sP{R}{-}{l} \\
 &    - \sP{L}{+}{u} - \sP{L}{-}{u} - \sP{R}{+}{u} - \sP{R}{-}{u} \\
 & \leq \sP{L}{-}{l} - \sP{R}{-}{u} \\
 & \leq \sA{L}{\reals}{\flipX_1 - a} - \sA{R}{\reals}{\flipX_1 + a}.
\end{align*}

$\tilde{f}_L(\flipX_1) - \tilde{f}_L(\flipX_1) - c < \sA{\reals}{\reals}{\flipX_1} + \sA{L}{\reals}{\flipX_1 - a} + \sA{R}{\reals}{\flipX_1 + a}
$
as $-c \leq |c| < \sA{L}{\reals}{\flipX_1} \leq \sA{\reals}{\reals}{\flipX_1}$. 
\end{iproof_of}

\newpage
\SingleLayerLimit* 
\begin{iproof_of}{thm:ub}
\label{prf:SingleLayerLimit}
Suppose $\alpha \in (0,1]$, and assume, for the sake of contradiction, 
that $f$ is a single-layer \relu-network with weights $N,M$ and biases $b,d$ that provably $\alpha$-robustly 
classifies $\lceil \frac{2}{\alpha}\rceil + 5$ flips.

We begin the proof by labeling the intermediate states of interval analysis for a point $\flipX_j$ with interval 
radius $\alpha$ of the network $f$:
\begin{align*}
v^{-}_{\alpha,j} & \defeq \relu(N\flipX_j + b - |N|\alpha) \\
v^{+}_{\alpha,j} & \defeq \relu(N\flipX_j + b + |N|\alpha) \\
w_{\alpha,j}     & \defeq v^+_{j,\alpha} - v^{-}_{j,\alpha} \\
c_{\alpha,j}     & \defeq v^+_{j,\alpha} + v^{-}_{j,\alpha} \\
f^\#(\langle \flipX_j, \alpha \rangle) & \defeq \langle \frac{1}{2}Mc_{\alpha,j} + d, \frac{1}{2}|M|w_{\alpha,j} \rangle,
\end{align*}
where the notation $|X|$ means the point-wise absolute value (i.e., $|X|_i = |X_i|$).

Then our assumption for contradiction tells us that for any $j \in [k]$ we know that $(-1)^j (M c_{\alpha,j} + 2d) \geq |M| w_{\alpha,j}.$

Let $j$ be such that \cref{lemma:gen_lb} tells us 
has $\flipL_j \alpha^{-1}( f(\flipX_j + \alpha) + f(\flipX_j - \alpha)) < \sA{\reals}{\reals}{\flipX_j} + \sA{R}{\reals}{\flipX_j - \alpha} + \sA{L}{\reals}{\flipX_j + \alpha}$.

We note that by expanding the definitions, sums, and meaning of absolute value,
we can derive that $Mc_{\alpha,j} + 2d = M(v^+_{\alpha,j} + v^+_{\alpha,j}) + 2d = f(\flipX_j + \alpha) + f(\flipX_j - \alpha)$.
so our assumption for contradiction thus implies 
$|M| w_{\alpha,j} \leq (-1)^j (M c_{\alpha,j} + 2d) = (-1)^j(f(\flipX_j+\alpha) + f(\flipX_j - \alpha)$.

We perform the following deduction:
\begin{align*}
 (-1)^j(f(\flipX_j+\alpha) + f(\flipX_j - \alpha)) 
& \geq  \sum_{N_i \flipX_j + b_i \geq -\alpha|N_i|} |M_i|(N_i x + b_i + \alpha|N_i|) - |M|v^{-}_{\alpha,j} \\
& \geq  \sum_{N_i \flipX_j + b_i \geq 0} |M_i|(N_i x + b_i + \alpha|N_i|) - |M|v^{-}_{\alpha,j} \\
& \geq  |M|\relu(Nx + b) + \sum_{N_i \flipX_j + b_i \geq 0} \alpha |M_iN_i| \\
& - \sum_{N_i \flipX_j + b_i \geq \alpha|N_i|} |M_i|(N_i x + b_i - \alpha|N_i|) \\
& \geq  \sum_{N_i \flipX_j + b_i \geq 0} \alpha |M_iN_i| + \sum_{N_i \flipX_j + b_i \geq \alpha|N_i|} \alpha |M_iN_i| \\
& \geq  \alpha \left( \sum_{N_i \flipX_j + b_i \geq 0} |M_iN_i| + \sum_{N_i \flipX_j + b_i \geq \alpha|N_i|} |M_iN_i| \right) \\
& \geq  \alpha ( \sA{\reals}{\reals}{\flipX_j} + \sA{R}{\reals}{\flipX_j - \alpha} + \sA{L}{\reals}{\flipX_j + \alpha}) \\
& \geq  \flipL_j (f(\flipX_j + \alpha) + f(\flipX_j - \alpha))
\end{align*}
which is a contradiction.
\end{iproof_of}

\newpage
\section{Extended Proofs for General Impossibility Results}
\label{appendix:impossible}
Here we restate the theorems and show the full proofs for the results in \cref{sec:impossible}.

\subsection{Proofs for Relative Interior Lemmas}
\label{appendix:relint}

In the following lemmas, let $A,A',B,B',C$ be bounded and non-empty subsets of $\mathbb{R}^d$:

\ProjLemma*
\begin{iproof_of}{lemma:rel:proj}
Let $y \in A|_i$.  
Then there is some $x \in A$ such that $x_i = y$.
Then $x \in \relint(B)$ by $A\sqsubset B$.
Then there is some $\epsilon > 0$ such that $N_\epsilon(x) \cap \aff(B) \subseteq B$.
Then $N_\epsilon(x)|_i \cap \aff(B)|_i \subseteq B|_i$.
We know $\aff(B|_i) \subseteq \aff(B)|_i$: 
   given $z \in \aff(B|_i)$, it must be an affine combination of the $i$'th dimension of elements of $B$.
   Letting $z'$ be the same affine combination of those elements, $z'_i = z$, so $z \in \aff(B)|_i$.
Then $N_\epsilon(x|_i) \cap \aff(B|_i) \subseteq B|_i$ and thus $y \in \relint(B|_i)$.
\end{iproof_of}

\CartesianLemma*
\begin{iproof_of}{lemma:rel:cartesian}
Let $(x,x') \in A \times A'$.
Then because $x \in A$ we know $x \in \relint(B)$  and respectively $x' \in \relint(B')$.
Then there is some $\epsilon > 0$ such that $N_\epsilon(x) \cap \aff(B) \subseteq B$, 
and $\epsilon' > 0$ such that $N_{\epsilon'}(x') \cap \aff(B') \subseteq B'$.
We know $(A \cap A') \times (B \cap B') \subseteq (A\times B) \cap (A' \times B')$:
  $(a,b) \in (A \cap A') \times (B \cap B')$ implies $a \in A\cap A'$ and $b \in B\cap B'$, so 
  $(a,b) \in A \times B$ and $(a,b) \in A' \times B'$ so $(a,b) \in (A\times B) \cap (A' \times B')$.

Then, we know $\aff(B \times B') \subseteq \aff(B) \times \aff(B')$:
   $(b,b') \in \aff(B \times B')$ implies $(b,b')$ is an affine combination of elements of $B \times B'$ which 
   implies $b$ is an affine combination of elements from $B$ and $b'$ is an affine combination of elements from $B'$, so $(b,b') \in \aff(B) \times \aff(B')$.

Thus for $\lambda = \min\{\epsilon, \epsilon'\}$ we know $N_\lambda(x,x') \cap \aff(B \times B') \subseteq B \times B'$.
Thus $(x,x') \in \relint(B \times B')$.
\end{iproof_of}

\DownUnionLemma*
\begin{iproof_of}{lemma:rel:down_union}
Suppose $x \in A \cup B$. Then $x \in A$ or $x\in B$.  Either way, we know $x \in \relint(C)$.
\end{iproof_of}

\DownHullLemma*
\begin{iproof_of}{lemma:rel:down_hull}
$C \in \balls^d$ implies $C|_i \in \balls$, and thus $C|_i$ is convex so $\relint(C|_i)$ is convex.
We know $A|_i \sqsubset C|_i$ by \cref{lemma:rel:proj}, so $\bHull(A|_i) \sqsubset C|_i$ by convexity of $C|_i$ 
and that $\bHull$ is the convex hull in one dimension. Thus, by \cref{lemma:rel:cartesian}, we know 
$\bHull(A|_1)\times \cdots \times \bHull(A|_d) \sqsubset C|_1 \times \cdots \times C|_d$.  
Because $C\in \balls^d$ we know $C = C|_1 \times \cdots \times C|_d$ and similarly that $\bHull(A) = \bHull(A|_1)\times \cdots \times \bHull(A|_d)$.
Thus, $\bHull(A) \sqsubset C$.
\end{iproof_of}

\newpage
\subsection{Proofs for Inversion and Impossibility Theorems}
\label{appendix:imprecision}

\InversionLemma*
\begin{iproof_of}{lemma:inversion}
  \label{prf:InversionLemma}
  The proof is by structural induction on the construction of the network $f$, assuming the lemma itself as the
  induction hypothesis for any network with fewer operations than $f$.  
  First, assume $Y \sqsubset \bHull(f(X))$.

    \StrIndCase{Sequential Computation}{$f = g \circ h$.}
    \begin{subproof}
      By definition, $Y \sqsubset \bHull(g \circ h(X))$.
      Thus, there exists some $H \sqsubset \bHull(h(X))$ 
        such that $Y \subseteq g^\#(H)$ by the induction hypothesis on $g$.  
      Applying the induction hypothesis again with the network $h$, 
        we get a set $X'\sqsubset \bHull(X)$ such that $H \subseteq h^\#(X')$.
      Thus, $Y \subseteq g(H) \subseteq g^\#\circ h^\#(X') = f^\#(X')$.
    \end{subproof}

    \StrIndCase{Relational Duplication}{$f(x) = (x, x)$.}
    \begin{subproof}
      Then $Y|_1 \sqsubset \bHull(X)$ and $Y|_2 \sqsubset \bHull(X)$ by
        \cref{lemma:rel:proj},
      We choose $X' = \bHull(Y|_1 \cup Y|_2)$ which we know by
        \cref{lemma:rel:down_union},
        is such that $X' \sqsubset \bHull(X)$.
      Thus, $Y|_1 \subseteq X'$ and $Y|_2 \subseteq X'$, so $Y \subseteq X'\times X' = f^\#(X')$.
    \end{subproof}

    \StrIndCase{Non-Relational Parallel Computation}{$f(x_1,x_2) = (g_1(x_1), g_2(x_2))$.}
    \begin{subproof}
      Then $Y|_1 \sqsubset \bHull(g_1(X|_1))$ and $Y|_2 \sqsubset \bHull(g_1(X|_1))$ 
        by definition.
      Then applying the induction hypothesis twice produces $L \sqsubset \bHull(X|_1)$ 
        and $R \sqsubset \bHull(X|_2)$ such that 
        $Y|_1 \subseteq g^\#_1(L)$ and $Y|_2 \subseteq g^\#_1(R)$.
      Then we choose $X' = L \times R$ which we know by
         \cref{lemma:rel:cartesian},
         is such that $X' \sqsubset \bHull(X)$.
      Then $Y \subseteq g^\#_1(X'|_1) \times g^\#_2(X'|_2) = f^\#(X')$.
    \end{subproof}

    \StrIndCase{Constant}{$f(x) = c$.}
    \begin{subproof}
      Here, any subset $X'\sqsubset X$ will suffice.  
      Then we can let $X'=\{\bcenter{X}\}$.
    \end{subproof}

    \StrIndCase{Multiplication by a Constant}{$f(x) = c\cdot x$ for $c\neq 0$.}
    \begin{subproof}
      Let $X' = \ball{|c^{-1}| \bradius{Y}}{ c^{-1} \bcenter{Y} }$.
      Then clearly, $Y \subseteq f(X')$.  
      It remains to show that $X' \sqsubseteq \bHull(X)$.
      For the remainder of this subproof, because we know that $d=1$ we will write $x_l = \inf{X}$, $x_u = \sup{X}$, 
      $y_l = \inf{Y}$ and $y_u = \sup{Y}$.  
      We note that $x_l = \bcenter{\bHull(X)} - \bradius{\bHull(X)}$ and so on.
      Supposing $c > 0$ (the other case is analogous) and $x_l < x_u$ (the proof is similar when they are equal),
      we have $c x_l < y_l \leq y_u < c x_u$ by $Y \sqsubset \bHull(f(X))$.  

      Then we know $x_l < |c^{-1}|y_l \leq |c^{-1}|y_u < x_u$, and thus $X' \sqsubset \bHull(X)$.
    \end{subproof}

    \StrIndCase{Activation}{$f(x) = \relu(x)$.}
    \begin{subproof}
      Again, because we know that $d=1$ we will write $x_l \defeq \inf{X}$, $x_u \defeq \sup{X}$, and $y_l \defeq \inf{Y}$ and $y_u \defeq \sup{Y}$.
      We know that $\inf{\bHull(f(X))} = \relu(x_l)$ and $\sup{\bHull(f(X))} = \relu(x_u)$.
      Thus by $Y \sqsubset \bHull(f(X))$ we know $\relu(x_l) \leq y_l \leq y_u \leq \relu(x_u)$.
      We then have two cases we need to address:
      
      \prfcase{Suppose}{$x_u > 0$.}
      \begin{subproof}
        Here we define $X' = [y_l,y_u]$.  
        We thus have $x_l \leq \relu(x_l) < y_l \leq y_u < \relu(x_u) = x_u$ provided $x_l < x_u$.  
        Otherwise we know $x_l = \relu(x_l) = y_l = y_u = \relu(x_u) = x_u$ so we have $X' \sqsubset \bHull(X)$.
      \end{subproof}
      
      \prfcase{Suppose}{$x_u \leq 0$.}
      \begin{subproof}
        Define $X' = \{ \frac{x_u + x_l}{2} \}$.
        We know $\relu(m) = 0 = y_l = y_u$ and thus $X' \sqsubset \bHull(X)$.
      \end{subproof}

      Thus, in both cases we can find $X' \sqsubset \bHull(X)$ such that $Y \subseteq f(X') \subseteq f^\#(X')$
    \end{subproof}

    \StrIndCase{Addition}{$f(x_1,x_2) = x_1 + x_2$.}
    \begin{subproof}
      Conveniently again, $Y$ is one-dimensional.  Either $\bHull(f(X))$ is a single point or it is not:

      \prfcase{Assume}{$\bHull(f(X))$ is a single point.}
      \begin{subproof}
          Then $\inf{Y} = \sup{Y} = \inf{\bHull(f(X))} = \sup{\bHull(f(X))}$. 
          Let $y = \inf{Y}$.
          In this case, we know there is some compact and non-empty set $Z\subseteq \reals$ such that 
          $X = \{(x', y - x') \given x' \in Z\}$.  Then we can pick $X' = \{(\bcenter{\bHull(Z)},y - \bcenter{\bHull(Z)})\}$ 
          which is the singleton-set containing the center of the \boxhull{} of $X$  and thus $X' \sqsubset \bHull(X)$ by \cref{lemma:rel:center}.
      \end{subproof}

      \prfcase{Otherwise}{$\bHull(f(X))$ is not a single point.}
      \begin{subproof}
          We know $\inf{\bHull(f(X))} < \inf{Y} \leq \sup{Y} < \sup{\bHull(f(X))}$.
          Because $Y$ is one-dimensional and a relative subset of the non-singular $\bHull(f(X))$ we know $Y \sqsubset \bHull(f(\bHull(X)))$.

          Let $a,b,r_a,r_b, y,r_y$ be as follows:
          \[
            \begin{aligned}
            a &= \bcenter{\bHull(X|_1)}, & r_a &= \bradius{\bHull(X|_1)}, \\
            b &= \bcenter{\bHull(X|_2)}, & r_b &= \bradius{\bHull(X|_2)}, \\
            y & = \bcenter{Y},           &\text{and }\; r_y &= \bradius{Y}.
            \end{aligned}
          \]
          Then $\bHull(X) = \ball{r_a}{a} \times \ball{r_b}{b}$
          and thus $Y \sqsubset f(\ball{r_a}{a} \times \ball{r_b}{b}) = \ball{r_a+r_b}{a + b}$.

          Choose $X' = \ball{r}{x}$ for $x$ and $r$ defined as:
          \[
          \begin{aligned}
            x_1 &= a + r_a \frac{y - a - b}{r_a + r_b},  & r_1 &= \frac{r_y r_a}{r_a + r_b},  \\ 
            x_2 &= b + r_b \frac{y - a - b}{r_a + r_b}, &\text{and }\; r_2 &= \frac{r_y r_b}{r_a + r_b}. 
          \end{aligned}
          \]
          Then clearly, $x_1 + x_2 = y$, and $r_1 + r_2 = r_y$ so $Y \subseteq f(X')$.

          Thus $r_y < r_a + r_b$ by $Y \sqsubset B_{r_a + r_b}(a+b)$. 

          This also tells us that $y + r_y < a + b + r_a + r_b$ and $y - r_y > a + b - r_a - r_b$.
          If $r_a \neq 0$ we can derive $x_1 + r_1 < a + r_a$ and $x_1 - r_1 > a - r_a$.
          Similarly, if $r_b \neq 0$ we can derive $x_2 + r_2 < b + r_b$ and $x_2 - r_2 > b - r_b$.
          Thus, $X_1 \sqsubset \ball{r_a}{a}$ and $X_2 \sqsubset \ball{r_b}{b}$, 
          and thus by \cref{lemma:rel:cartesian} we have
          $X' \sqsubset \ball{r_a}{a} \times \ball{r_b}{b} = \bHull(X)$.
        \end{subproof}
      Thus, in both cases, there exists an $X' \sqsubset \bHull(X)$ such that $Y \subseteq f(X') \subseteq f^\#(X')$.
    \end{subproof}
  As any feed forward neural network (without input-value dependent loops)
can be expressed using these operations without modifying the result under interval analysis, by induction 
$\exists X' \sqsubset \bHull(X) \st  Y \subseteq f^\#(X')$.
\end{iproof_of}

\newpage
\ImprecisionThm*
\begin{iproof_of}{theorem:imprecision}
\label{prf:ImprecisionThm}
  The proof is by structural induction on the construction of the network $f$, assuming the theorem itself as the
  induction hypothesis for any network with fewer operations than $f$.
  
  Let $f \from \mathbb{R}^n \to \mathbb{R}^m$ be a feed forward network with \relu{} activations, and let $x \in \mathbb{R}^m$
  and let $N \subseteq f^{-1}(x)$ be compact and non-empty.
  Then $f$ is one of the following cases:

    \StrIndCase{Sequential Computation}{$f = g \circ h$}
    \begin{subproof}
      We first know that $N \subseteq h^{-1} \circ g^{-1}(x)$ by the definition of $f$.
      We then infer that $h(N) \subseteq \bHull(N)$ is compact and non-empty by application of the continuous function $h$.
      Thus, by induction on $h(N)$ and $g$ there is some $M' \sqsubset \bHull(h(N))$ such that $x \in g^\#(M')$.
      Thus, by \cref{lemma:inversion}, we know that there is some $M\sqsubset \bHull(N)$ such that $M' \subseteq h^\#(M)$.
      Thus, $x \in g^\# \circ h^\#(M) = f^\#(M)$.
    \end{subproof}

    \StrIndCase{Relational Duplication}{$f(y) = (y, y)$.}
    \begin{subproof}
      We have $N\subseteq \{x_1\} \cap \{x_2\}$ so $N=\{x_1\}$. By singleton reflexivity, $N \sqsubset \bHull(N)$.  
      Thus, $x \in f^\#(N)$.
    \end{subproof}

    \StrIndCase{Non-Relational Parallel Computation}{$f(x_1,x_2) = (g_1(x_1), g_2(x_2))$.}
    \begin{subproof}
      First we know $N|_1 \subseteq g_1^{-1}(x_1)$ and $N|_2 \subseteq g_2^{-1}(x_2)$ 
      by projection and that $N|_1$ and $N|_2$ are still compact and non-empty.
      Thus, by the induction hypothesis twice we see that there are boxes
      $M_1 \sqsubset \bHull(N|_1)$ and $M_2 \sqsubset \bHull(N|_2)$ 
      such that 
      $x_1 \in g_1^\#(M_1)$ and
      $x_2 \in g_2^\#(M_2)$.
      Then $M_1 \times M_2 \sqsubset \bHull(N)$ by 
        \cref{lemma:rel:cartesian}.
      Then $x_1 \in g_1^\#(M_1 \times M_2)$ 
      and $x_2 \in g_2^\#(M_1 \times M_2)$ 
      by soundness.
      Thus, there is some box $M\sqsubset \bHull(N)$ such that $x \in f^\#(M)$.
    \end{subproof}

    \StrIndCase{Constant}{$f(y) = c$.} 
    \begin{subproof}
      We know $x = c$ and thus $f^{-1} = \mathbb{R}$.
      If we let $M=\{\bcenter{\bHull(N)} \} \sqsubset \bHull(N)$,
      then $x \in f^\#(M)$.
    \end{subproof}

    \StrIndCase{Multiplication by a Constant}{$f(y) = c\cdot y$ for $c\neq 0$.}
    \begin{subproof}
      We know $f^{-1}(x) = \{ c^{-1} \cdot x \} = N \sqsubset \bHull(N)$ by $N$ being non-empty 
      and thus $x \in f^\#(N)$.
    \end{subproof}

    \StrIndCase{Activation}{$f(y) = \relu(y)$.}
    \begin{subproof}
      Then $x=\relu(y)$ can either be zero or greater than zero.

      \prfcase{Case}{$x > 0$}
      \begin{subproof}
        $N = \{x\}$ by def. of \relu, and we know $\{x\} \sqsubset \bHull(N)$
        and $x \in f^\#(\{x\})$.
      \end{subproof}

      \prfcase{Case}{$x = 0$}
      \begin{subproof}
       $N = (-\infty, 0]$ by def. of \relu. 
       Thus $\{\bcenter{\bHull(N)}\} \sqsubset \bHull(N)$
       and $x \in f^\#(\{\bcenter{\bHull(N)}\})$.
      \end{subproof}

      Because $x$ is the result of a \relu, it must have been one of these two possibilities, 
      and in both cases we could find some $M \sqsubset \bHull(N)$ such that $x \in f^\#(M)$.
    \end{subproof}

    \StrIndCase{Addition}{$f(y_1,y_2) = y_1 + y_2$.}
    \begin{subproof}
      In this case, we know $N \subseteq f^{-1}(x) = \{ (a, x - a) \given a \in \mathbb{R}\}$.
      We pick $M = \{ \bcenter{\bHull(N)} \} \sqsubset \bHull(N)$.
      Given $N$ is bounded, we know:
      \begin{align*}
        \bcenter{\bHull(N)} &= \left(\frac{\inf N|_1 +\sup N|_1}{2} ,\frac{\inf N|_2 +\sup N|_2}{2}\right) \\
                            &= \left(\frac{\inf N|_1 +\sup N|_1}{2} ,\frac{2x - \inf N|_1  - \sup N|_1}{2}\right).
      \end{align*}
      We can rewrite $f(\bcenter{\bHull(N)})$ as
      \[
        f(\bcenter{\bHull(N)}) = \frac{\inf N|_1 +\sup N|_1}{2} + \frac{2x - \inf N|_1  - \sup N|_1}{2} = x.
      \]
      Thus, $x = f(\bcenter{\bHull(N)}) \in f^\#(M)$
    \end{subproof}

  Thus, $\exists M \sqsubset \bHull(N) \ldot x \in f^\#(M).$
\end{iproof_of}

\end{document}